\documentclass{article} % For LaTeX2e
\usepackage[preprint]{colm2026_conference}

\usepackage{float}
\usepackage{caption}
\usepackage{microtype}
\usepackage{hyperref}
\usepackage[utf8]{inputenc}
\usepackage{url}
\usepackage{booktabs}
\usepackage{graphicx}
\usepackage{multirow}
\usepackage[utf8]{inputenc}
\usepackage{tabularx}
\usepackage{amsmath}
\usepackage{amssymb}
\usepackage{xcolor}
\usepackage{enumitem}
\usepackage{pifont}
\usepackage{needspace}

\newcommand{\cmark}{\ding{51}}
\newcommand{\xmark}{\ding{55}}

\usepackage{lineno}

\definecolor{darkblue}{rgb}{0, 0, 0.5}
\hypersetup{colorlinks=true, citecolor=darkblue, linkcolor=darkblue, urlcolor=darkblue}

\title{LUDOBENCH: Evaluating LLM Behavioural Decision-Making Through Spot-Based Board Game Scenarios in Ludo}

\author{Ojas Jain, Dhruv Kumar \\
Department of Computer Science and Information Systems, BITS Pilani \\
\texttt{\{f20210976,dhruv.kumar\}@pilani.bits-pilani.ac.in}
}

\begin{document}

\ifcolmsubmission
\linenumbers
\fi

\maketitle

\begin{abstract}
We introduce \textsc{LudoBench}, a benchmark for evaluating LLM strategic reasoning in Ludo, a stochastic multi-agent board game whose dice mechanics, piece capture, safe-square navigation, and home-path progression introduce meaningful planning complexity. \textsc{LudoBench} comprises 480 handcrafted spot scenarios across 12 behaviorally distinct decision categories, each isolating a specific strategic choice. We additionally contribute a fully functional 4-player Ludo simulator supporting Random, Heuristic, Game-Theory, and LLM agents. The game-theory agent uses Expectiminimax search with depth-limited lookahead to provide a principled strategic ceiling beyond greedy heuristics. Evaluating six models spanning four model families, we find that all models agree with the game-theory baseline only 40--46\% of the time. Models split into distinct behavioral archetypes: finishers that complete pieces but neglect development, and builders that develop but never finish. Each archetype captures only half of the game theory strategy. Models also display measurable behavioral shifts under history-conditioned grudge framing on identical board states, revealing prompt-sensitivity as a key vulnerability. \textsc{LudoBench} provides a lightweight and interpretable framework for benchmarking LLM strategic reasoning under uncertainty. All code, the spot dataset (480 entries) and model outputs are available at \url{https://anonymous.4open.science/r/LudoBench-5CBF/}
\end{abstract}

\section{Introduction}
Large Language Models have progressed from structured NLP benchmarks to cognitively demanding tasks including mathematical reasoning and commonsense inference \citep{Wang2018, Hendrycks2020, Cobbe2021, Talmor2018}. Models such as GPT-4, Llama-3, and DeepSeek achieve near-human performance on academic benchmarks \citep{OpenAI2023, Meta2024, Team2024}, yet a critical question persists: can LLMs reason strategically, planning ahead, weighing competing objectives, and deciding under uncertainty, or does their reasoning collapse under novel conditions? Game-playing environments, where rules are unambiguous and outcomes are verifiable, offer a uniquely controlled setting to investigate this question \citep{Kosinski2023, Hu2024, Xu2024}.

While chess, poker, and Go have received extensive AI research attention, Ludo has, to our knowledge, no existing benchmark or published LLM evaluation. Ludo is one of the world's most widely played board games, yet this gap persists despite its unique combination of game-theoretic properties. It is simultaneously stochastic (dice-driven), multi-agent (2--4 players), multi-piece\footnote{Throughout this paper, ``piece'' refers to Ludo game pieces. ``Token'' appears only in code variables and the LLM prompt template, where it denotes game pieces, not LLM tokens.} (four independent pieces requiring coordination), and complete-information. This configuration is absent from most existing LLM game benchmarks. Existing benchmarks cover deterministic games (chess, Go), incomplete-information games (poker), and cooperative games (Diplomacy), but no benchmark addresses stochastic race games where agents manage multiple assets under probabilistic transitions. \textsc{LudoBench} adds this missing category to the game-based evaluation landscape (Table~\ref{tab:positioning}).

We ask: to what extent can LLMs make strategically sound decisions in a stochastic multi-agent game? The LLM receives a natural-language board state and must output a piece index with justification \citep{Zhuang2025, Huang2024}. This formulation simultaneously tests spatial reasoning and strategic preference across competing objectives.

We introduce \textsc{LudoBench}, a spot-based evaluation framework that uses Ludo to measure LLM strategic reasoning across 12 decision categories. Each \emph{spot} is a handcrafted board configuration that forces a specific strategic choice computable by a heuristic oracle. Five persona conditions (\texttt{none}, \texttt{aggressive}, \texttt{greedy}, \texttt{safe}, \texttt{unforgiving}) enable systematic instruction-following analysis. A paired grudge design presents identical boards with neutral versus conflict narratives, isolating history-conditioned framing effects from board-state effects. Our main contributions are as follows:
\begin{itemize}
    \item \textbf{\textsc{LudoBench} Benchmark and Dataset:} A spot-based evaluation framework for LLM strategic reasoning in a stochastic multi-agent board game, comprising 480 engine-validated scenarios across 12 decision categories with ground-truth legal moves and history-conditioned grudge testing.

    \item \textbf{Ludo Simulator:} A fully functional 4-player simulation environment supporting three agent types (\texttt{RandomAgent}, \texttt{HeuristicAgent}, \texttt{LLMAgent}) with configurable matchup settings.

    \item \textbf{Game-Theory Baseline:} A chance-aware search agent implementing Expectiminimax recursion for two-player settings and Expectimax-MaxN for three and four-player settings, with depth-limited lookahead. This agent provides a game-theory reference for Ludo spot evaluation, enabling measurement of how far LLM decisions deviate from principled strategic play.
\item \textbf{Empirical Profiles:} We evaluate six models spanning five families across all 12 categories and 5 persona conditions, producing detailed per-model behavioral profiles. These enable systematic comparison of strategic tendencies, rule compliance, persona responsiveness, and history sensitivity across model families and scales.
\end{itemize}
\begin{table}[t]
\centering
\scriptsize
\setlength{\tabcolsep}{4pt}
\renewcommand{\arraystretch}{1.05}
\caption{Positioning of \textsc{LudoBench} against existing LLM game benchmarks. \cmark\,=\,present, \xmark\,=\,absent.}
\label{tab:positioning}
\vspace{0.3em}
\begin{tabular}{@{}lccccc@{}}
\toprule
 & \textbf{Chess} & \textbf{Go} & \textbf{Poker} & \textbf{Dipl.} & \textbf{Ludo} \\
\textbf{Property} & \citep{Campbell2002} & \citep{Silver2016} & \citep{Zhuang2025} & \citep{FAIR2022} & (ours) \\
\midrule
Stochastic            & \xmark & \xmark & \cmark & \xmark & \cmark \\
Complete information   & \cmark & \cmark & \xmark & \cmark & \cmark \\
Multi-agent (3+)       & \xmark & \xmark & \cmark & \cmark & \cmark \\
Multi-piece coord.     & \cmark & \cmark & \xmark & \xmark & \cmark \\
Identical pieces       & \xmark & \cmark & \xmark & \xmark & \cmark \\
% Legal moves withheld   & \xmark & \xmark & \xmark & \xmark & \cmark \\
Spot-based eval.       & \xmark & \xmark & \cmark & \xmark & \cmark \\
Behavioral metrics     & \xmark & \xmark & \xmark & \xmark & \cmark \\
History-cond.\ testing & \xmark & \xmark & \xmark & \xmark & \cmark \\
Persona evaluation     & \xmark & \xmark & \xmark & \xmark & \cmark \\
\bottomrule
\end{tabular}
\end{table}
\section{Related Work}
\label{sec:related}
AI game-playing has progressed from game-specific engines such as Deep Blue in chess \citep{Campbell2002}, AlphaGo in Go \citep{Silver2016}, and Libratus in poker \citep{Brown2018, Brown2019} to LLM-based approaches that use language as the game interface. Early work showed that LLMs can navigate text-based environments like Zork \citep{Cote2019}, but studies on chess exposed limits in multi-step planning and positional evaluation \citep{Xu2024, Ruoss2024}. Agent frameworks such as ReAct \citep{Yao2023react} introduced iterative game loops with self-correction, and work on Werewolf \citep{Xu2024} and Diplomacy demonstrated rudimentary theory-of-mind, though with poor long-horizon planning. A critical limitation unifies this prior work: nearly all evaluated domains feature deterministic mechanics or simple stochastic structure. Stochastic race games, which require probabilistic reasoning over dice outcomes, multi-piece coordination, and opponent modeling, remain unexamined. \textsc{LudoBench} addresses this gap through Ludo's dice-driven stochasticity, four pieces per player, and competing capture-versus-progress incentives.

Several benchmarks systematize LLM game evaluation. \citet{costarelli2024gamebench} standardizes win-rate metrics but aggregates results without scenario-level breakdowns. \citet{Zhuang2025} introduced spot-based evaluation with 11,000 poker scenarios. However, existing benchmarks omit stochastic board games, provide legal moves directly to the model, and do not measure behavioral tendencies such as risk aversion or history-conditioned decision shifts (cf.\ Table~\ref{tab:positioning}). \textsc{LudoBench} extends the spot-based paradigm to a multi-agent stochastic setting with per-category metrics (\texttt{capture\_rate}, \texttt{safe\_rate}, \texttt{change\_rate}) that decompose performance into interpretable decision-level signals.

Chain-of-Thought and Tree-of-Thought prompting have improved LLM reasoning \citep{Wei2022, Yao2023}, yet systematic biases such as positional bias, risk aversion, and framing sensitivity persist and have rarely been examined where optimal actions are computable at scale. The heuristic oracle and paired grudge scenarios in \textsc{LudoBench} enable reproducible measurement of LLM behavioral biases against a deterministic reference.

\section{Background: The Ludo Environment}
\label{sec:background}
Ludo is one of the world's most widely played board games, with hundreds of millions of players across South Asia, Europe, and Africa, yet to our knowledge, no prior work has studied LLM decision-making in its domain. Prior AI research on Ludo has focused on reinforcement learning agents, including TD($\lambda$) and Q-learning players \citep{Alhajry2012, Chhabra2015, Tubaishat2023}, but none has evaluated LLMs as game agents or proposed a behavioral benchmarking framework. This absence is surprising given Ludo's unique combination of game-theoretic properties. It is simultaneously \emph{stochastic} (dice rolls introduce genuine probabilistic uncertainty at every turn), \emph{multi-agent} (2--4 competing players), \emph{multi-piece} (each player controls four identical pieces at different stages of progress, requiring parallel coordination), and \emph{complete-information} (all positions are visible to all players). This configuration is absent from every existing LLM game benchmark. Unlike chess, where each piece has a unique movement rule and captures are deterministic, Ludo requires coordinating four functionally identical pieces whose optimal advancement depends on dice probabilities rather than positional tactics. Unlike poker, where the agent manages a single hidden resource and must reason about incomplete information, Ludo presents complete information but demands simultaneous management of multiple independent assets across a shared board. Ludo's core cognitive challenge is thus deciding \emph{which copy of the same piece to advance} under dice-driven uncertainty, testing whether LLMs can reason about risk distribution, portfolio management, and parallel goal pursuit simultaneously. The board layout is illustrated in Figure~\ref{fig:ludo_board}. Full board zone specifications are provided in Appendix~\ref{app:board_model}.

Existing LLM game benchmarks span deterministic perfect-information games (chess, Go), incomplete-information games (poker), and cooperative games (Diplomacy), but omit stochastic race games entirely. This leaves a gap in the evaluation of LLM strategic reasoning across diverse game-theoretic environments. \textsc{LudoBench} fills this missing category by adding stochastic multi-piece coordination to the benchmark landscape (Table~\ref{tab:positioning}). Ludo's global cultural ubiquity further strengthens its value as a testbed, as LLMs have almost certainly encountered Ludo's rules during pretraining. Finally, Ludo's small action space (at most four pieces) enables exhaustive behavioral profiling per decision point while remaining strategically rich enough to reveal preference hierarchies across 12 categories and 5 personas.
\section{Methodology}
\label{sec:method}

\paragraph{Overview.}
% \paragraph{Overview.}
We propose a behavioral evaluation framework that tests LLM decision-making in Ludo through spot-based evaluation, which presents isolated single-decision board states. The spot mode is the primary contribution of this work. It removes game trajectory as a confound and measures one behavioral tendency at a time. 

\paragraph{Game Environment.}\label{sec:method_env}
The Ludo engine encodes all game rules and serves as both the game substrate and the legal-move oracle. At each decision point, the full board state is passed to the LLM as a numerical piece-position list. Legal moves are intentionally withheld. Instead, the model receives high-level game rules and must determine which moves are valid from these descriptions. Invalid outputs thus become a meaningful compliance signal.

\paragraph{Spot Scenario Dataset.}\label{sec:method_spots}
Each spot is a handcrafted board state that guarantees exactly one strategic tradeoff, with choices computable by a game-theory oracle. The 480 scenarios (40 per category) span 12 categories across four behavioral dimensions: \textbf{Aggression} (capture versus alternatives), \textbf{Safety} (safe-square preference), \textbf{Progress and Compliance} (home entry, overshoot avoidance, extra-turn management), and \textbf{History Sensitivity} (paired grudge and neutral conditions on identical boards). The grudge design is a key methodological contribution. The same board state is presented with neutral and grudge-framing histories, ensuring that any decision shift is attributable purely to narrative context. Table~\ref{tab:dataset_distribution} (Appendix) confirms balanced coverage across 2-, 3-, and 4-player configurations. Full per-category definitions are in Appendix~\ref{app:categories}. 
\paragraph{LLM Agent and Output Processing.}\label{sec:method_agent}
The prompt contains the board state, full rules, dice value, and optionally a persona instruction (\texttt{none}, \texttt{aggressive}, \texttt{greedy}, \texttt{safe}, \texttt{unforgiving}) or a history paragraph. The model responds in a strict format: \texttt{<piece\_index> | <reason>}. Responses are validated against engine-computed legal moves. Invalid outputs are flagged and replaced by a random legal move for continuity. Invalid rates are recorded \emph{pre-fallback}, and all behavioral rates are computed on valid outputs only. This design keeps compliance and decision quality as separate measurement strata.

\paragraph{Comprehensive Evaluation.}\label{sec:method_baselines}
We compare LLM agents against three baselines defining a skill ladder of increasing strategic sophistication. The \textbf{Random Agent} selects uniformly from legal moves and defines the performance floor. The \textbf{Heuristic Agent} applies a deterministic one-step scoring policy that uses board progress as the base score, a $+100$ capture bonus, and a $+20$ safe-square bonus, always selecting the highest-scoring move. It provides a greedy and interpretable reference beyond chance. The full scoring formula is in Table~\ref{tab:app_heuristic} (Appendix~\ref{app:heuristic}). The \textbf{Game-Theory Agent} performs depth-limited Expectiminimax search (depth $h{=}2$; see Appendix~\ref{app:gt_agent}) for two-player games and Expectimax-MaxN search for three- and four-player games. It models opponent best responses and averages over all six dice outcomes at chance nodes. At the depth cutoff, a weighted linear evaluator scores board progress, finished pieces, base-piece penalty, and safe-square count. Unlike the heuristic, this agent reasons multiple moves ahead under uncertainty, providing a principled strategic ceiling. Full algorithmic details are in Appendix~\ref{app:gt_agent}. Figure~\ref{fig:winrate_matrix} confirms the skill ladder through head-to-head play. Table~\ref{tab:design_choices} summarizes key design decisions.

\begin{figure*}[t]
  \centering
  \begin{minipage}[c]{0.48\textwidth}
    \centering
    \includegraphics[width=0.85\textwidth]{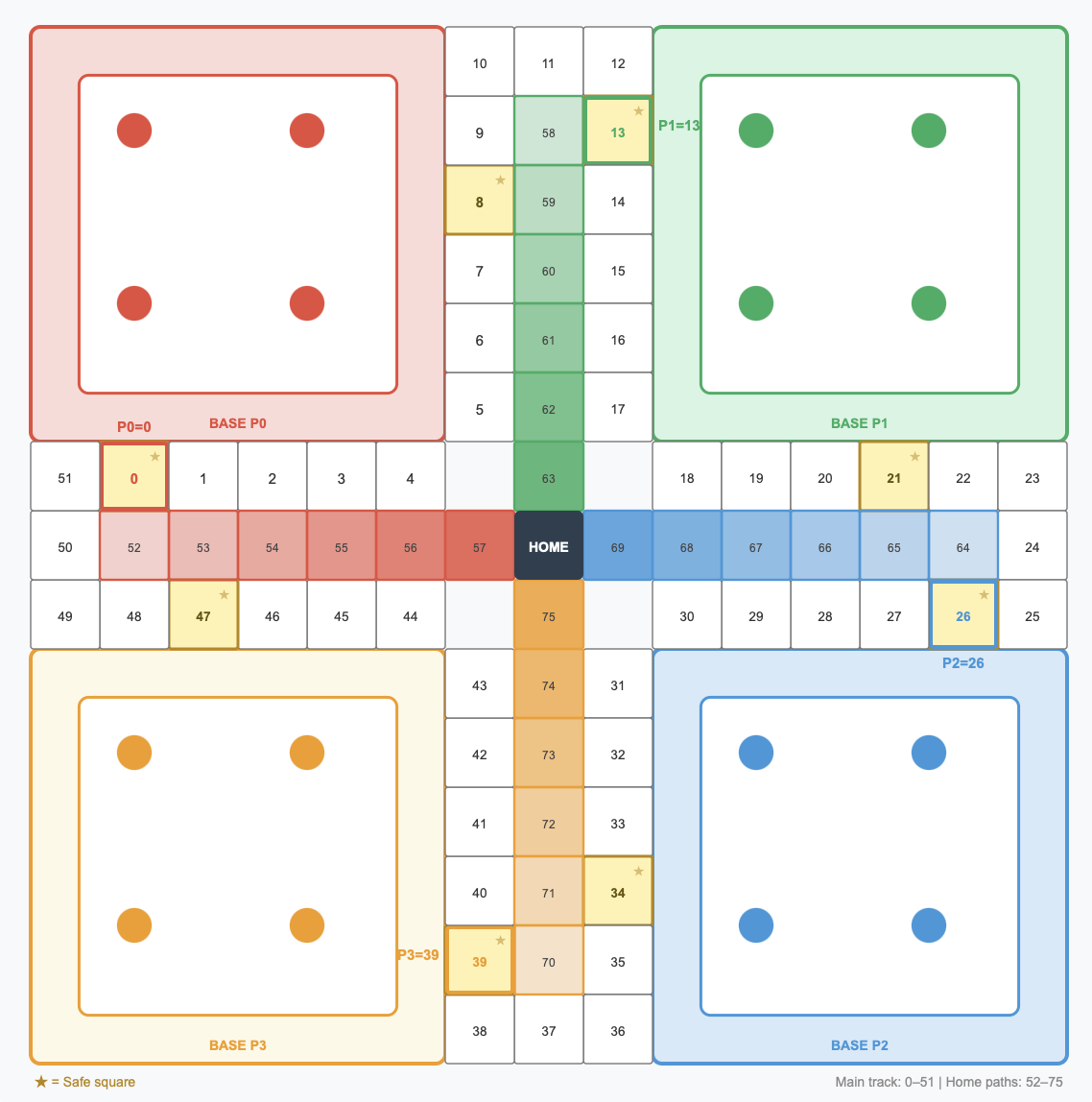}
    \captionof{figure}{\textbf{Annotated Ludo Board.} Main track (0--51), colored home paths, safe squares, and base areas. Player start positions are P0\,=\,0, P1\,=\,13, P2\,=\,26, P3\,=\,39.}
    \label{fig:ludo_board}
  \end{minipage}
  \hfill
  \begin{minipage}[c]{0.48\textwidth}
    \centering
    \includegraphics[width=\textwidth]{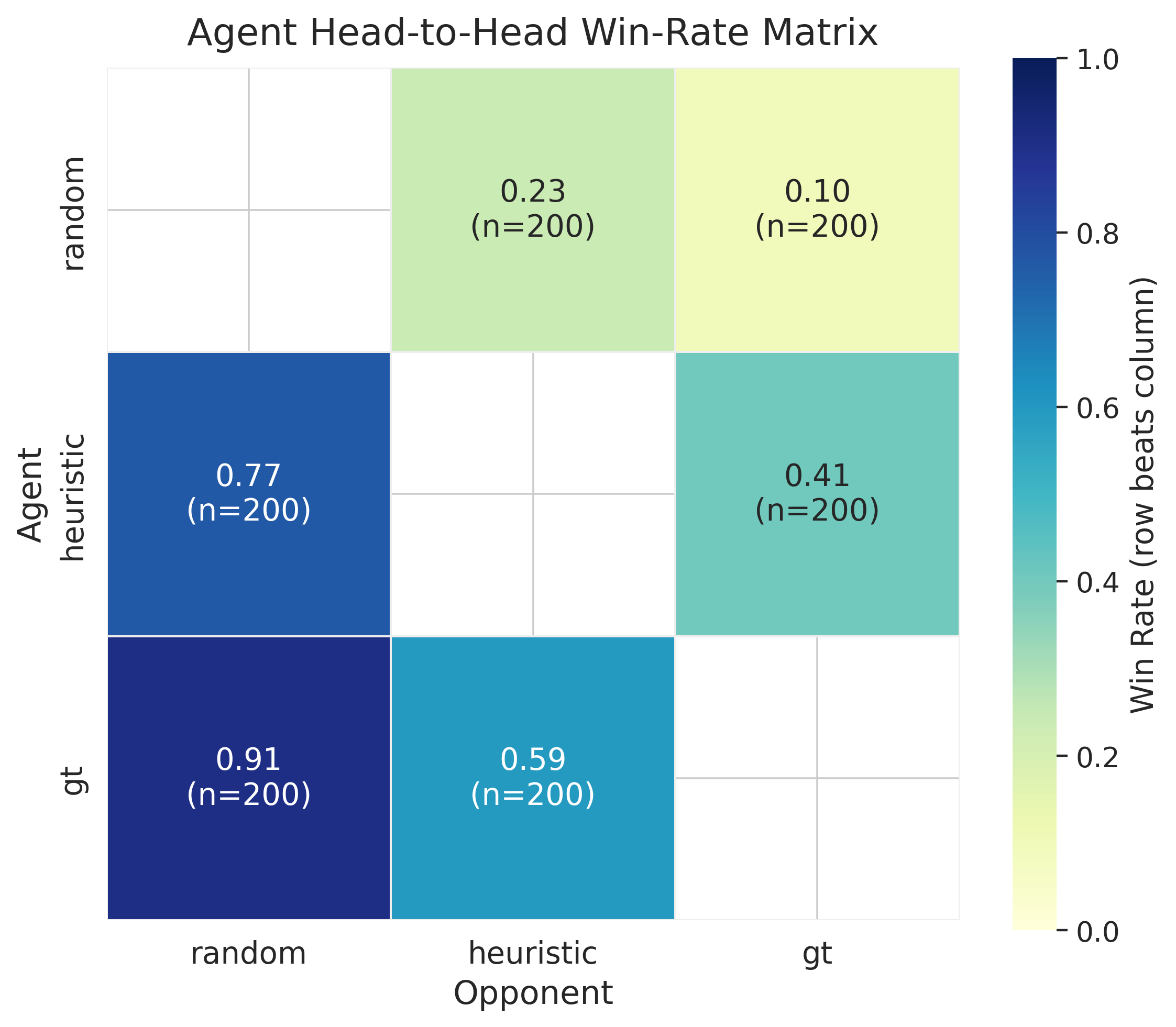}
    \captionof{figure}{\textbf{Agent Head-to-Head Win-Rate Matrix.} Win rates over 200 games confirm the baseline skill ladder: Random $<$ Heuristic $<$ GT. GT's 59\% vs.\ Heuristic validates depth-limited search advantage.}
    \label{fig:winrate_matrix}
  \end{minipage}
\end{figure*}

\section{Experimental Setup}
\label{sec:setup}

Table~\ref{tab:setup} summarizes the evaluation configuration. All models are accessed via their respective APIs in a zero-shot setting with temperature$=$0 and no system prompt beyond the spot template (Appendix~\ref{app:prompt}). Each of the six models is evaluated on all 480 spots under each of the 5 persona conditions, yielding 2,400 spot evaluations per model (14,400 total). For the \texttt{grudge\_paired} category, each of the 40 board-state pairs is evaluated under all 5 personas, producing 400 paired comparisons per model. Behavioral metrics are computed on valid outputs only; invalid rates are recorded pre-fallback.
\begin{table}[h]
  \centering
  \begin{minipage}[t]{0.45\textwidth}
    \centering
    \scriptsize
    \setlength{\tabcolsep}{3pt}
    \renewcommand{\arraystretch}{1.1}
    \caption{Evaluation configuration.}
    \label{tab:setup}
    \vspace{0.3em}
    \begin{tabular}{@{}ll@{}}
      \toprule
      \textbf{Parameter} & \textbf{Value} \\
      \midrule
      Models         & QP, Q7B, DS, Haiku, Scout, Gemma \\
      Families       & Qwen, DeepSeek, Anthropic, Meta, Google \\
      Categories     & 12 (40 each; 80 grudge) \\
      Total spots    & 480 boards  \\
      Personas       & 5 (none, aggr., greedy, safe, unforg.) \\
      Evals/model    & 2,400 (480 $\times$ 5) \\
      Total evals    & 14,400 (6 $\times$ 2,400) \\
      Temperature    & 0 (deterministic) \\
      Legal moves    & Withheld from prompt \\
      \bottomrule
    \end{tabular}
  \end{minipage}
  \hfill
  \begin{minipage}[t]{0.52\textwidth}
    \centering
    \scriptsize
    \setlength{\tabcolsep}{3pt}
    \renewcommand{\arraystretch}{1.1}
    \caption{Key design choices and justifications.}
    \label{tab:design_choices}
    \vspace{0.3em}
    \begin{tabular}{@{}lp{3.5cm}@{}}
      \toprule
      \textbf{Design Choice} & \textbf{Justification} \\
      \midrule
      Handcrafted spots      & Guarantees the target tradeoff is present \\
      Paired grudge design   & Isolates narrative effects from board-state effects \\
      Strict integer output  & Enables automated validity checking at scale \\
      Invalid pre-fallback   & Separates compliance from decision quality \\
      Game-theory baseline   & Provides a strategic ceiling beyond greedy heuristics \\
      \bottomrule
    \end{tabular}
  \end{minipage}
\end{table}
\section{Results and Analysis}
\label{sec:results}

We evaluate six LLMs on all 12 spot categories without any persona instruction (the neutral setting) and compare their decisions against our Game-Theory(GT) baseline. The six models are Qwen-Plus (QP), Qwen-2.5-7B-Instruct (Q7B), DeepSeek-Chat (DS), Claude-3.5-Haiku (Haiku), Llama-4-Scout (Scout), and Gemma-3-12B-IT (Gemma).

Our evaluation reveals three key findings. First, following the rules does not mean playing well. Two models with near-perfect rule compliance make completely opposite strategic choices. Second, every LLM falls into one of two behavioral camps: ``finishers'' who complete pieces, and ``builders'' who develop new ones. The GT agent does \emph{both}, and no LLM comes close to matching it. Third, all models agree with the GT agent only about 40--46\% of the time, meaning that more than half of their moves are strategically suboptimal.

\subsection{Rule Compliance}
\label{sec:invalid}

\begin{figure*}[t]
  \centering
  \begin{minipage}[c]{0.48\textwidth}
    \centering
    \includegraphics[width=\textwidth]{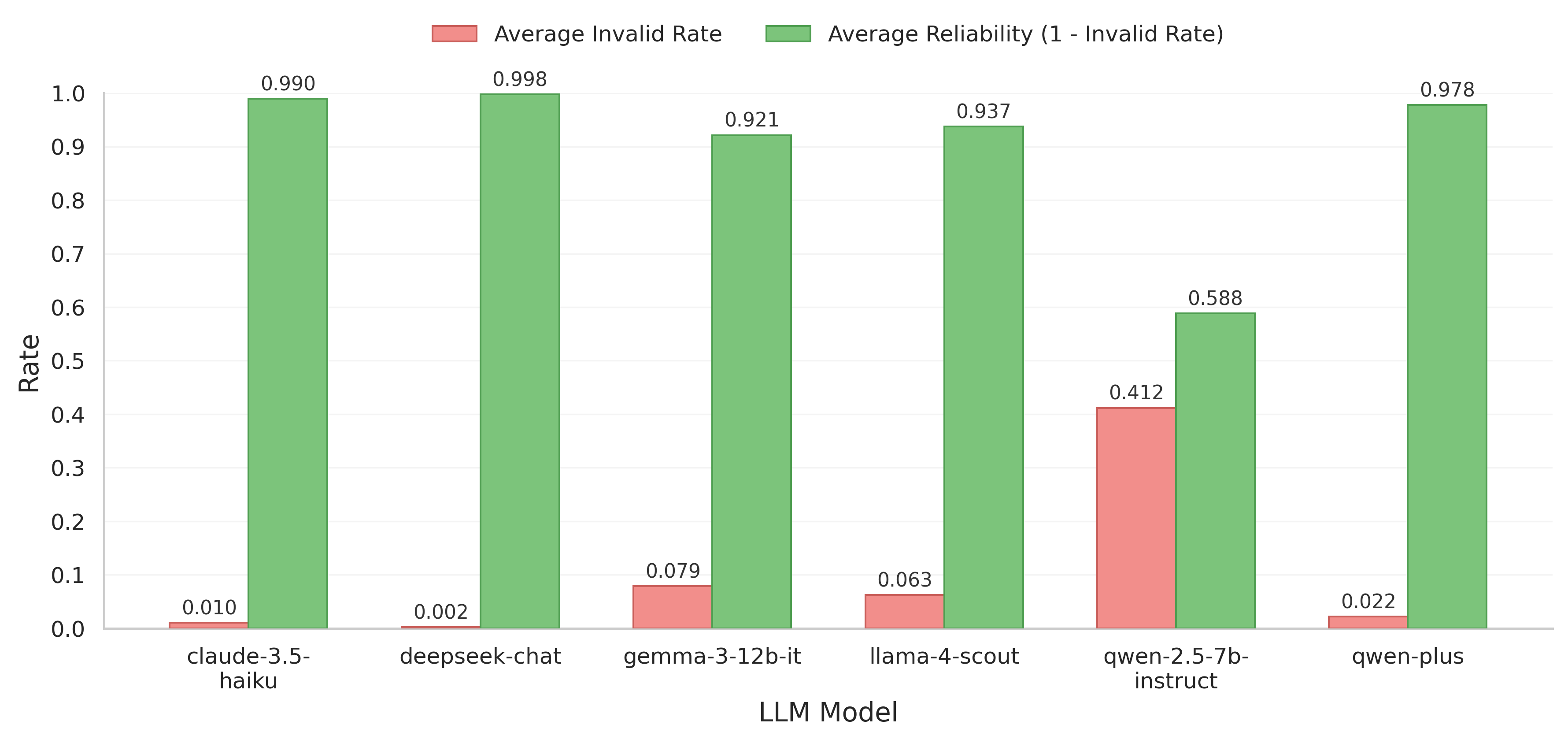}
    \captionof{figure}{\textbf{Average Invalid Rate and Reliability} (1 $-$ invalid rate) across all non-grudge categories. Three compliance tiers are visually evident.}
    \label{fig:invalid_reliability}
  \end{minipage}
  \hfill
  \begin{minipage}[c]{0.48\textwidth}
    \centering
    \includegraphics[width=\textwidth]{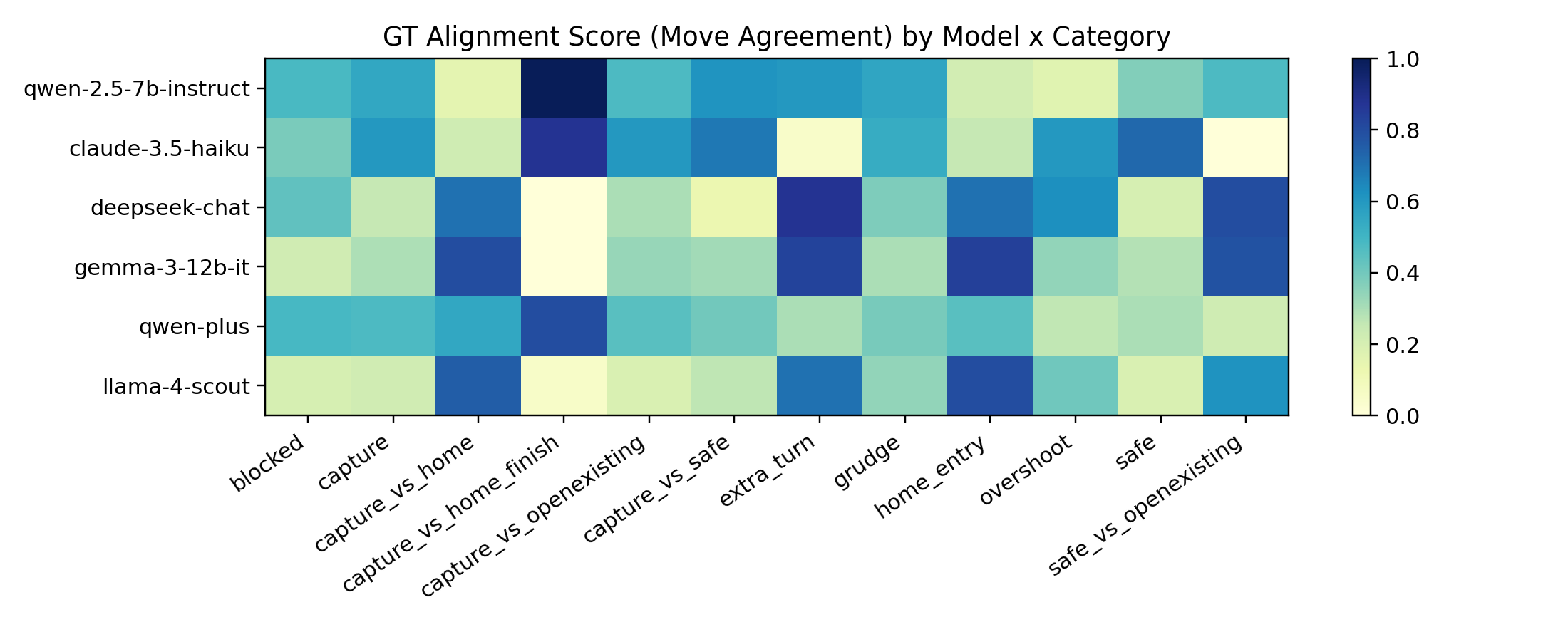}
    \captionof{figure}{\textbf{GT Alignment Score} (move agreement with GT agent) by model and category. Dark = high agreement; light = systematic disagreement.}
    \label{fig:gt_heatmap}
  \end{minipage}
\end{figure*}

 Figure~\ref{fig:invalid_reliability} summarizes the illegal-move rates across models. The full per-category breakdown is provided in Table~\ref{tab:invalid} (Appendix~\ref{app:metrics}). The models fall into three clear groups. \textbf{Tier~1 (near-perfect):} DeepSeek-Chat and Claude-3.5-Haiku almost never make illegal moves, with both falling below 1\%. These models reliably determine legal moves from the provided rule descriptions. \textbf{Tier~2 (mostly reliable):} Qwen-Plus (4\%), Llama-4-Scout (8\%), and Gemma-3-12B-IT (10\%) produce valid moves in most cases but struggle in specific situations. For example, Qwen-Plus fails 43\% of the time on overshoot scenarios, where the dice roll would take a piece past the exact finish square. \textbf{Tier~3 (unreliable):} Qwen-2.5-7B makes illegal moves approximately 40\% of the time, making its strategic behavior difficult to interpret. Notably, even Q7B scores 0\% invalid on the three dice$=$6 categories, where the choice is simply between bringing out a new piece or moving an existing one. This suggests that simpler decisions are easier for weaker models.

Rule compliance does not simply depend on model size. Gemma-3-12B-IT (10\% invalid) performs worse than the much larger Qwen-Plus (4\%), suggesting that training methodology matters as much as scale. All metrics are formally defined in Table~\ref{tab:metrics} (Appendix~\ref{app:metrics}).
\begin{table*}[t]
\centering
\small
\begin{tabular}{@{}lcccccc@{}}
\toprule
\textbf{Category} & \textbf{QP} & \textbf{Q7B} & \textbf{DS} & \textbf{Haiku} & \textbf{Scout} & \textbf{Gemma} \\
\midrule
\texttt{blocked}                   & 0.03 & 0.28 & 0.03 & 0.03 & 0.13 & 0.20 \\
\texttt{capture\_vs\_home\_finish} & 0.00 & 0.60 & 0.00 & 0.00 & 0.08 & 0.13 \\
\texttt{capture\_vs\_home}         & 0.00 & 0.68 & 0.00 & 0.00 & 0.10 & 0.13 \\
\texttt{capture\_vs\_openexisting} & 0.00 & 0.00 & 0.00 & 0.00 & 0.08 & 0.03 \\
\texttt{capture\_vs\_safe}         & 0.00 & 0.55 & 0.00 & 0.05 & 0.15 & 0.20 \\
\texttt{capture}                   & 0.00 & 0.50 & 0.00 & 0.00 & 0.10 & 0.15 \\
\texttt{overshoot}                 & 0.43 & 0.55 & 0.00 & 0.00 & 0.08 & 0.05 \\
\texttt{safe}                      & 0.00 & 0.53 & 0.00 & 0.00 & 0.08 & 0.03 \\
\texttt{safe\_vs\_openexisting}    & 0.00 & 0.00 & 0.00 & 0.00 & 0.03 & 0.08 \\
\texttt{extra\_turn}               & 0.00 & 0.00 & 0.00 & 0.00 & 0.00 & 0.00 \\
\texttt{home\_entry}               & 0.00 & 0.65 & 0.00 & 0.00 & 0.13 & 0.08 \\
\midrule
\textbf{Mean (excl.\ grudge)}      & \textbf{0.04} & \textbf{0.39} & \textbf{$<$0.01} & \textbf{$<$0.01} & \textbf{0.08} & \textbf{0.10} \\
\bottomrule
\end{tabular}
\caption{Illegal-move rate (\texttt{llm\_invalid\_rate}) by model and category under the neutral persona. Each value is the fraction of spots where the model's output was not a legal move.}
\label{tab:invalid}
\end{table*}

\subsection{Strategic Alignment with GT Agent}
\label{sec:gt_alignment}

To measure how \emph{strategically good} each model's decisions are, we compare them directly against the GT agent. For each of the 480 board positions, we check whether the LLM selects the same piece as the GT agent. We refer to this as the \textbf{GT alignment score}, which represents the fraction of decisions where a model makes the same move as the strongest baseline. Figure~\ref{fig:gt_heatmap} shows the full breakdown by model and category.

We observe that \textbf{all six models agree with the GT agent only 40--46\% of the time}. More than half of every model's decisions are strategically suboptimal, regardless of model family, size, or rule-compliance tier. No single model stands out as clearly superior.

However, the average hides an important pattern. When examined at the level of individual categories, alignment ranges from 0\% to 100\%. Qwen-2.5-7B agrees with the GT agent 100\% of the time in \texttt{capture\_vs\_home\_finish}, where it always finishes like the GT, but only 17\% in \texttt{overshoot} due to its high illegal-move rate. DeepSeek-Chat agrees 88\% in \texttt{extra\_turn}, where it favors developing new pieces in alignment with the GT, but 0\% in \texttt{capture\_vs\_home\_finish}, where it always captures instead of finishing. Claude-3.5-Haiku exhibits the inverse pattern, achieving 88\% alignment on finishing but only 5\% on development.

Each model is therefore ``right'' in categories where its natural playing style happens to match the GT, and ``wrong'' everywhere else. This indicates that these models have acquired \emph{parts} of a sound strategy from their training data, but none of them have learned the complete strategic picture.

\subsection{Behavioral Archetypes}
\label{sec:profiles}

\begin{figure*}[t]
  \centering
  \begin{minipage}[c]{0.48\textwidth}
    \centering
    \includegraphics[width=\textwidth]{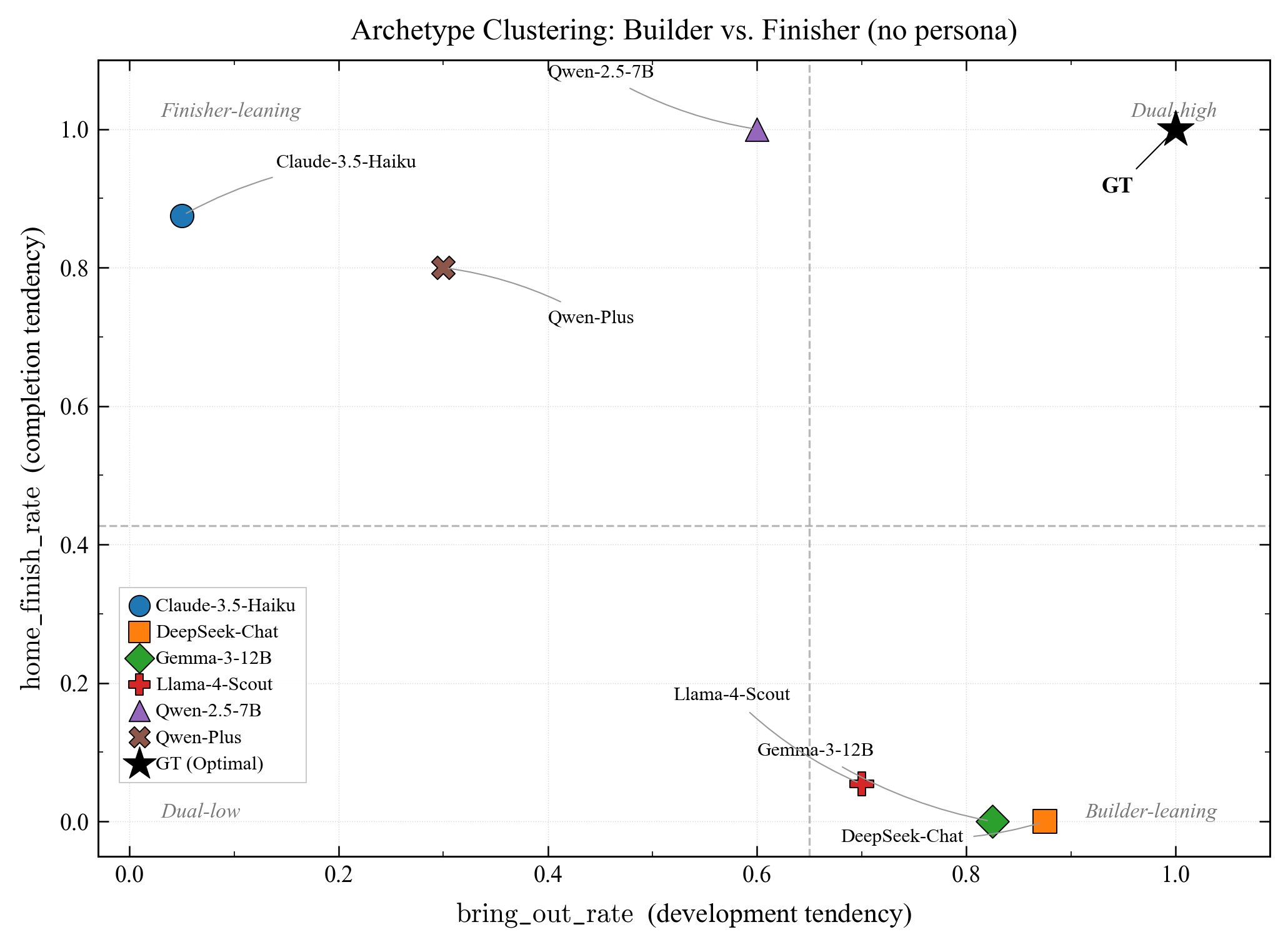}
    \captionof{figure}{\textbf{Archetype Clustering: Builder versus Finisher.} Each model is plotted by development tendency (x-axis) versus completion tendency (y-axis). The GT agent (star) exhibits both tendencies. LLMs split into finishers (top-left) and builders (bottom-right).}
    \label{fig:archetype_scatter}
  \end{minipage}
  \hfill
  \begin{minipage}[c]{0.48\textwidth}
    \centering
    \includegraphics[width=\textwidth]{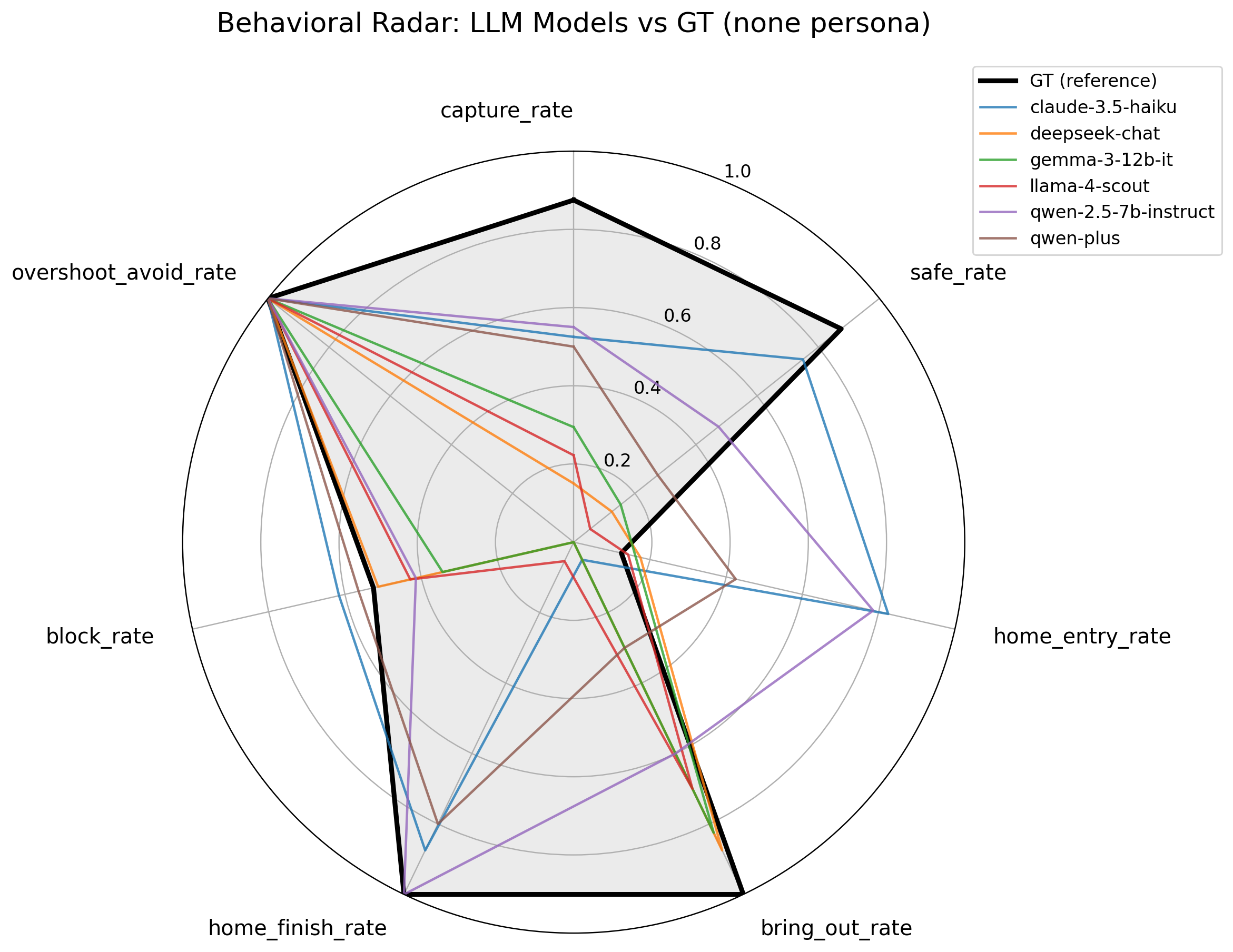}
    \captionof{figure}{\textbf{Behavioral Radar: LLMs versus GT Reference.} Each axis represents a behavioral metric. The GT agent (bold black) scores high on most axes. Each LLM matches the GT on some axes but falls short on others.}
    \label{fig:radar}
  \end{minipage}
\end{figure*}

The GT alignment patterns raise a natural question: what playing styles produce these results? Figure~\ref{fig:archetype_scatter} plots each model on two dimensions, namely how often it develops new pieces (x-axis) and how often it finishes pieces (y-axis). Full metrics are in Table 9 (Appendix~\ref{app:metrics}). The plot reveals a clear pattern: \textbf{models that develop new pieces do not finish them, and models that finish do not develop}. The GT agent sits alone in the top-right corner, as it always develops \emph{and} always finishes. No LLM approaches this balanced profile. The models split into two primary groups.

\textbf{Finishers} (Claude-3.5-Haiku, Qwen-2.5-7B, Qwen-Plus) focus on completing pieces. Haiku is the most extreme example. When it rolls a 6, it advances an existing piece 95\% of the time instead of bringing out a new one, whereas the GT always brings out a new piece. On the positive side, Haiku finishes pieces 88\% of the time when given the choice, which is close to the GT's 100\%. Finishers therefore get the ``finish when you can'' part of the strategy correct but miss the ``develop your board'' component.\\
\textbf{Builders} (DeepSeek-Chat, Gemma-3-12B-IT) focus on getting new pieces onto the board. DeepSeek-Chat brings out new pieces 88\% of the time, close to the GT's 100\%. However, when given a choice between finishing and capturing, both builders \emph{always} capture and \emph{never} finish, the exact opposite of the GT's preference. Finishing is a permanent step toward winning, whereas capturing only delays the opponent.\\
\textbf{Capture-oriented} (Llama-4-Scout) sits between the two groups with moderate development tendency (70\% bring-out rate) but a strong capture preference in tradeoff scenarios (95\% capture over finishing).Figure~\ref{fig:radar} overlays all profiles against the GT on seven behavioral axes, confirming that no LLM fills the GT's balanced shape.

\subsection{Tradeoff Analysis}
\label{sec:tradeoffs}

Some categories force the model to choose between two competing goals, such as capturing an opponent or finishing one of its own pieces. These categories are the most revealing, because the GT agent provides a clear indication of which choice is strategically superior. Figure~\ref{fig:tradeoff_bars} shows each model's preference as a diverging bar with the GT's choice marked as a green reference line, and full rates are provided in (Appendix~\ref{app:metrics}). The clearest finding comes from the \texttt{capture\_vs\_home\_finish} (CVF) category, where the GT agent \emph{always} finishes. Finishers align with this preference, with Q7B finishing 100\% of the time, Haiku 88\%, and QP 80\%, while builders consistently get this wrong, as DS and Gemma capture 100\% of the time and Scout captures 95\%. An interesting reversal appears in \texttt{capture\_vs\_home} (CVH) and \texttt{capture\_vs\_safe} (CVS), where the GT agent captures 98\% of the time. In these categories, finisher models get it wrong: Q7B captures only 15\% and Haiku only 20\% in CVH, preferring instead to enter the home path, while builder models are closer to the GT, with Gemma at 83\% and Scout at 78\%. This means that \textbf{each archetype gets only half the strategy right}: finishers are correct about finishing but wrong about capturing, while builders are wrong about finishing but right about capturing. In \texttt{capture\_vs\_openexisting} (CVO), the GT agent captures 68\% of the time, representing a more balanced tradeoff, and Qwen-Plus (62\%) comes closest to this reference point.

\subsection{Grudge Sensitivity}
\label{sec:grudge_results}

Our grudge scenarios test whether a model's decisions change when it is told that an opponent recently captured one of its pieces, even though the board state is exactly the same. The GT agent is completely unaffected by this framing, achieving a change\_rate of 0\%, and Table~\ref{tab:grudge} shows how the six LLMs respond to this manipulation. Qwen-Plus changes its decision 33\% of the time when told about a prior conflict, even on identical boards, and DeepSeek-Chat changes 20\% of the time, while the other four models are relatively stable with change rates between 0\% and 6\%. This is a concerning finding, as it demonstrates that simply adding a narrative about past events can shift strategic decisions without any change to the actual game state. One caveat is worth noting: some models, specifically Scout and Gemma, retaliate at 81--88\% regardless of narrative framing, which suggests that the retaliatory moves in our scenarios may also be strategically attractive on their own merits. The change\_rate metric, which measures \emph{any} shift in decision rather than only shifts toward retaliation, is therefore a cleaner measure of narrative sensitivity. By this measure, sensitivity is unrelated to both rule compliance and playing style, as Qwen-Plus, which has moderate compliance and a finisher archetype, is the most affected, while other finishers with better compliance show near-zero sensitivity.

\subsection{Persona Effects}
\label{sec:persona_effects}

\begin{figure*}[t]
  \centering
  \begin{minipage}[c]{0.48\textwidth}
    \centering
    \includegraphics[width=\textwidth]{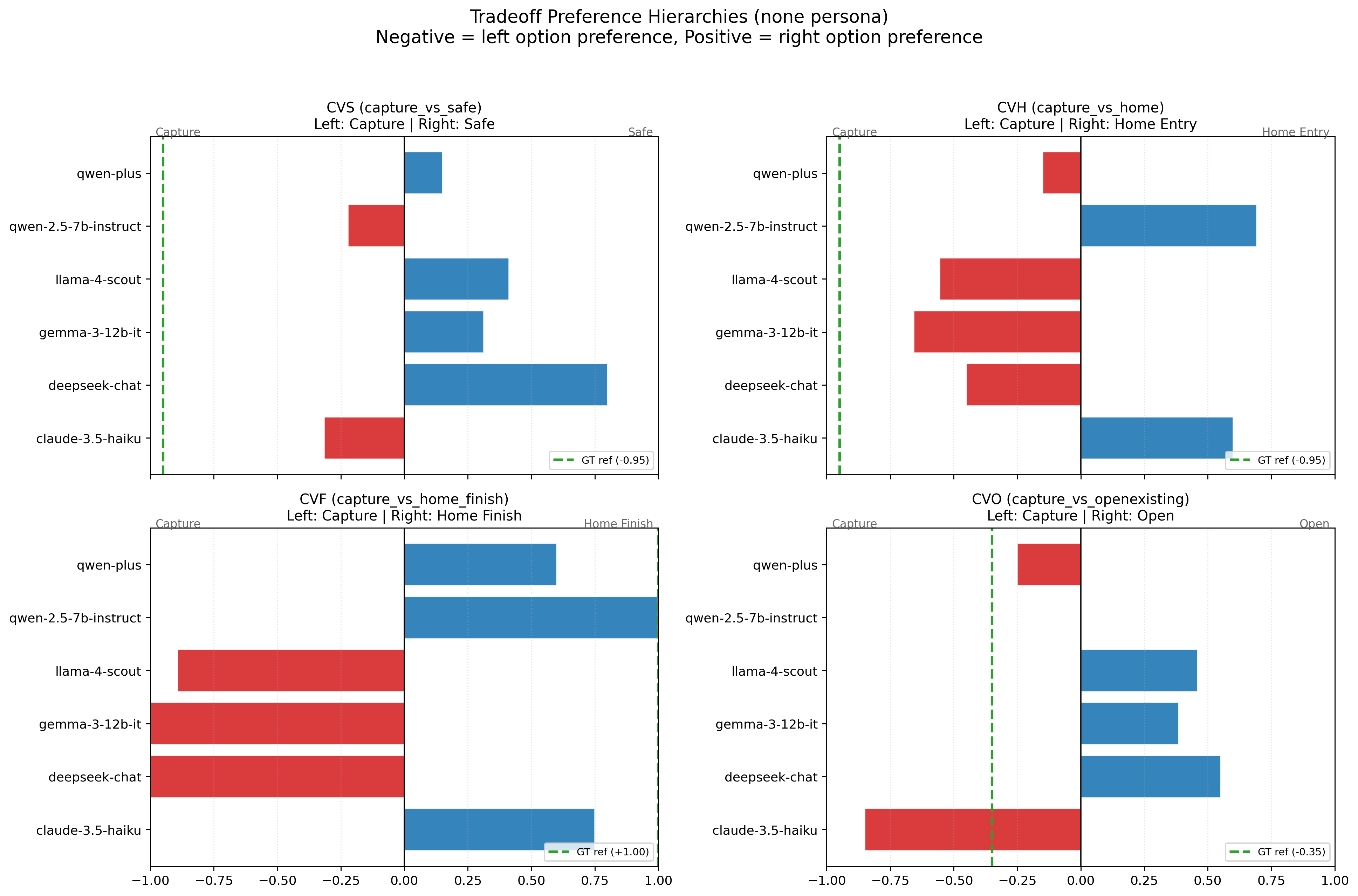}
    \captionof{figure}{\textbf{Tradeoff Preference Hierarchies.} Each bar shows a model's preference (left = capture, right = alternative). The green dashed line marks the GT choice. Bars on the same side as the line indicate correct decisions.}
    \label{fig:tradeoff_bars}
  \end{minipage}
  \hfill
  \begin{minipage}[c]{0.48\textwidth}
    \centering
    \includegraphics[width=\textwidth]{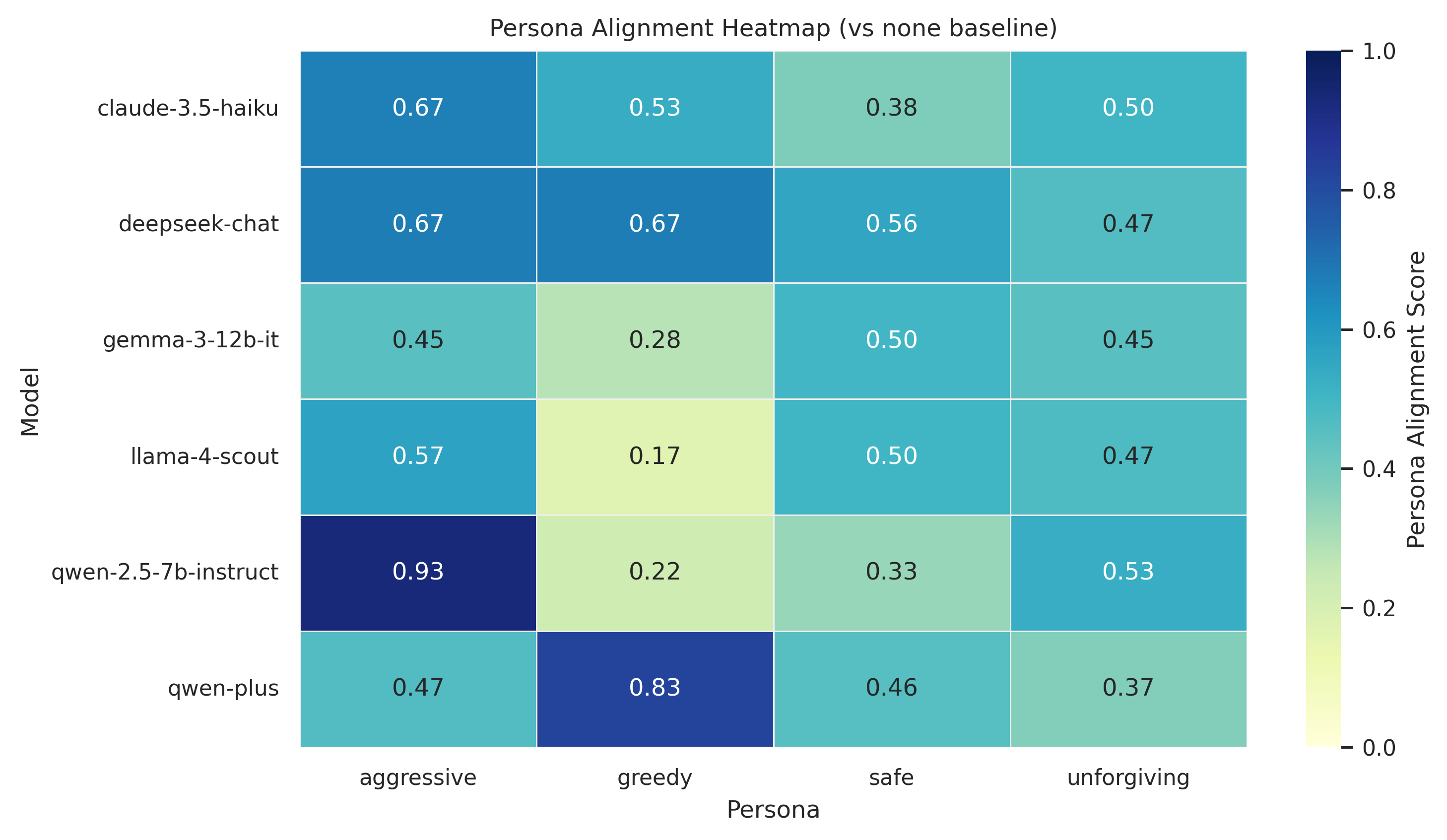}
    \captionof{figure}{\textbf{Persona Alignment Heatmap.} Each cell shows how strongly behavior shifts to match the persona (1.0 = perfect). Most scores fall between 0.3 and 0.5, indicating weak effects. Only Q7B-aggressive (0.93) and QP-greedy (0.83) show strong alignment.}
    \label{fig:persona_heatmap}
  \end{minipage}
\end{figure*}

We also test whether persona instructions (``play aggressively,'' ``play cautiously,'' etc.) shift behavior in the expected direction. Figure~\ref{fig:persona_heatmap} summarizes the results; full details are in Table~\ref{tab:persona_effects} (Appendix~\ref{app:persona_details}).
Persona instructions are generally weak and unreliable. Most alignment scores fall between 0.3 and 0.5. Only two combinations show strong effects, namely Q7B under the aggressive persona (0.93) and QP under the greedy persona (0.83). The greedy persona produces the strongest effects, likely because it specifies a concrete goal (``advance toward home'') rather than a vague behavioral style (``be aggressive''). Some instructions even produce the opposite of the intended effect. For example, instructing Haiku to ``play safely'' actually makes it \emph{more} aggressive in capture-versus-safe scenarios, raising its capture rate to 88\% from 66\% under the neutral condition.This paradoxical inversion suggests that persona instructions interact unpredictably with the model's pre-existing strategic biases rather than overriding them. These results highlight that persona-based steering is not a reliable mechanism for controlling LLM decision-making in strategic settings, and that developers should empirically verify the directional effects of any behavioral prompt before deployment.

\section{Discussion}
\label{sec:discussion}

Table~\ref{tab:summary} summarizes the key findings across all six models. Our findings have implications beyond Ludo. We highlight four takeaways for the broader LLM research community.
\textbf{Rule compliance does not imply strategic competence.} DeepSeek-Chat and Claude-3.5-Haiku both follow Ludo's rules almost perfectly, yet they adopt opposite strategies. Testing whether a model follows instructions correctly tells us nothing about how well the model reasons about tradeoffs.

\textbf{LLMs learn partial strategies, not complete ones.} The GT agent simultaneously develops new pieces and finishes them, and no LLM replicates this balanced approach. Each model has picked up one half of a good strategy from its training data, and the GT alignment scores (40--46\%) confirm that models agree with the GT only where their natural style matches.

\textbf{Narrative framing can shift decisions on identical board states.} Qwen-Plus changes its decision 33\% of the time when told about a prior conflict on exactly the same board. If game narratives can shift strategy, then conversational context in real deployments could similarly shift LLM decisions in hard-to-predict ways.

\textbf{Persona instructions are unreliable.} Most persona instructions have weak effects, and some produce the opposite of the intended behavior. Developers building LLM agents should use concrete and measurable instructions rather than vague behavioral prompts.
Table~\ref{tab:realworld} maps these findings to real-world deployment contexts.
\begin{table*}[t]
  \centering
  \begin{minipage}[t]{0.55\textwidth}
    \centering
    \scriptsize
    \setlength{\tabcolsep}{2pt}
    \caption{Cross-model comparison. GT alignment represents average move agreement with the GT agent. cvf/cvs alignment indicates whether the model's preferred choice matches the GT's action (\cmark\,=\,match, \xmark\,=\,opposite).}
    \label{tab:summary}
    \vspace{0.3em}
    \begin{tabular}{@{}lcccccc@{}}
      \toprule
      \textbf{Dimension} & \textbf{QP} & \textbf{Q7B} & \textbf{DS} & \textbf{Haiku} & \textbf{Scout} & \textbf{Gemma} \\
      \midrule
      Rule compl. & 4\% & 39\% & $<$1\% & $<$1\% & 8\% & 10\% \\
      Archetype & Bal.\ fin. & Goal-or. & Builder & Cons.\ fin. & Cap.-lean. & Builder \\
      GT align. & 43\% & 46\% & 46\% & 45\% & 40\% & 46\% \\
      cvf align. & \cmark & \cmark & \xmark & \cmark & \xmark & \xmark \\
      cvs align. & \xmark & \xmark & \xmark & partial & \xmark & \xmark \\
      Grudge chg. & 33\% & 0\% & 20\% & 5\% & 6\% & 3\% \\
      \bottomrule
    \end{tabular}
  \end{minipage}
  \hfill
  \begin{minipage}[t]{0.42\textwidth}
    \centering
    \scriptsize
    \setlength{\tabcolsep}{3pt}
    \caption{How \textsc{LudoBench} findings relate to real-world LLM deployment.}
    \label{tab:realworld}
    \vspace{0.3em}
    \begin{tabular}{@{}p{2.2cm}p{3.0cm}@{}}
      \toprule
      \textbf{Game Behavior} & \textbf{Real-World Parallel} \\
      \midrule
      Rule compliance & Legal rules, API constraints, regulations \\[4pt]
      Capture vs.\ safe & Risk in finance, medicine, autonomous driving \\[4pt]
      Develop vs.\ advance & Portfolio management, resource allocation \\[4pt]
      Narrative sensitivity & Prompt injection, adversarial context \\[4pt]
      Persona following & Agent alignment, role-based assistants \\
      \bottomrule
    \end{tabular}
  \end{minipage}
\end{table*}
\section{Conclusion}
\label{sec:conclusion}

We introduced \textsc{LudoBench}, a benchmark that evaluates LLM strategic reasoning through 480 handcrafted Ludo board scenarios across 12 decision categories. Evaluating six models from five families, we find that rule compliance does not predict strategic quality, that models split into finisher and builder archetypes each capturing only half of the GT agent's balanced strategy, and that narrative context can shift decisions on identical boards by up to 33\%. These findings suggest that standard win-rate evaluations may miss behavioral patterns that emerge more clearly from spot-level analysis. Future work could extend \textsc{LudoBench} to additional models, multi-step planning, end-to-end game simulations, and other game domains.
\section*{Limitations}
Our scenarios are hand-built and do not cover every possible Ludo decision; some categories (e.g., \texttt{safe\_vs\_openexisting}) showed ceiling effects where all models behaved identically. We use 40 spots per category due to API cost constraints, so our rates are point estimates without confidence intervals. Spot-based evaluation tests one decision at a time, not multi-step planning, which could be studied in future work. We tested six models from five families at temperature$=$0; results may differ for other models or settings. Ludo has a small action space (at most four pieces) and we use English-only prompts, limiting generalizability to more complex domains or other languages. Finally, our evaluation is entirely spot-based; end-to-end simulation where LLM agents play complete multi-turn games against each other or against baseline agents remains to be conducted, and would provide complementary insights into cumulative strategic performance over full game trajectories.

\section*{Acknowledgements}  
The authors acknowledge the use of AI tools such as ChatGPT, Claude, and Gemini for improving the presentation and grammar of this paper. All the results, analysis, and proposed techniques remain a concrete representation of the author's contributions. The authors take full responsibility for the contents in this paper.

\bibliographystyle{colm2026_conference}
\bibliography{colm2026_conference}

\begin{thebibliography}{26}
\providecommand{\natexlab}[1]{#1}
\providecommand{\url}[1]{\texttt{#1}}
\expandafter\ifx\csname urlstyle\endcsname\relax
  \providecommand{\doi}[1]{doi: #1}\else
  \providecommand{\doi}{doi: \begingroup \urlstyle{rm}\Url}\fi

\bibitem[Alhajry et~al.(2012)Alhajry, Alvi, and Ahmed]{Alhajry2012}
Majed Alhajry, Faisal Alvi, and Moataz Ahmed.
\newblock {TD($\lambda$) and Q-learning based Ludo players}.
\newblock In \emph{2012 IEEE Conference on Computational Intelligence and Games (CIG)}, pp.\  83--90. IEEE, 2012.
\newblock \doi{10.1109/CIG.2012.6374142}.

\bibitem[Brown \& Sandholm(2018)Brown and Sandholm]{Brown2018}
Noam Brown and Tuomas Sandholm.
\newblock Superhuman {AI} for heads-up no-limit poker: {L}ibratus beats top professionals.
\newblock \emph{Science}, 359\penalty0 (6374):\penalty0 418--424, 2018.

\bibitem[Brown \& Sandholm(2019)Brown and Sandholm]{Brown2019}
Noam Brown and Tuomas Sandholm.
\newblock Superhuman {AI} for multiplayer poker.
\newblock In \emph{Science}, volume 365, pp.\  885--890, 2019.

\bibitem[Campbell et~al.(2002)Campbell, Hoane~Jr, and Hsu]{Campbell2002}
Murray Campbell, A~Joseph Hoane~Jr, and Feng-hsiung Hsu.
\newblock Deep {B}lue.
\newblock \emph{Artificial Intelligence}, 134\penalty0 (1--2):\penalty0 57--83, 2002.

\bibitem[Chhabra \& Tomar(2015)Chhabra and Tomar]{Chhabra2015}
Veenus Chhabra and Kavita Tomar.
\newblock Artificial intelligence: Game techniques ludo---a case study.
\newblock \emph{ACSIT}, 2\penalty0 (6):\penalty0 549--553, 2015.

\bibitem[Cobbe et~al.(2021)Cobbe, Kosaraju, Bavarian, Chen, Jun, Kaiser, Plappert, Tworek, Hilton, Nakano, et~al.]{Cobbe2021}
Karl Cobbe, Vineet Kosaraju, Mohammad Bavarian, Mark Chen, Heewoo Jun, Lukasz Kaiser, Matthias Plappert, Jerry Tworek, Jacob Hilton, Reiichiro Nakano, et~al.
\newblock Training verifiers to solve math word problems.
\newblock \emph{arXiv preprint arXiv:2110.14168}, 2021.

\bibitem[Costarelli et~al.(2024)Costarelli, Allen, Hauksson, Sodunke, Hariharan, Cheng, Li, Clymer, and Yadav]{costarelli2024gamebench}
Anthony Costarelli, Mat Allen, Roman Hauksson, Grace Sodunke, Suhas Hariharan, Carlson Cheng, Wenjie Li, Joshua Clymer, and Arjun Yadav.
\newblock Gamebench: Evaluating strategic reasoning abilities of llm agents.
\newblock \emph{arXiv preprint arXiv:2406.06613}, 2024.

\bibitem[C\^{o}t\'{e} et~al.(2019)C\^{o}t\'{e}, K\'{a}d\'{a}r, Yuan, Kybartas, Barnes, Fine, Moore, Hausknecht, El~Asri, Adada, Tay, and Trischler]{Cote2019}
Marc-Alexandre C\^{o}t\'{e}, \'{A}kos K\'{a}d\'{a}r, Xingdi Yuan, Ben Kybartas, Tavian Barnes, Emery Fine, James Moore, Matthew Hausknecht, Layla El~Asri, Mahmoud Adada, Wendy Tay, and Adam Trischler.
\newblock {TextWorld}: A learning environment for text-based games.
\newblock \emph{arXiv preprint arXiv:1806.11532}, 2019.

\bibitem[{FAIR}(2022)]{FAIR2022}
{FAIR}.
\newblock Human-level play in the game of diplomacy by combining language models with strategic reasoning.
\newblock \emph{Science}, 378\penalty0 (6624):\penalty0 1067--1074, 2022.

\bibitem[{Google DeepMind}(2024)]{Team2024}
{Google DeepMind}.
\newblock {Gemma}: Open models based on {G}emini research and technology.
\newblock \emph{arXiv preprint arXiv:2403.08295}, 2024.

\bibitem[Hendrycks et~al.(2021)Hendrycks, Burns, Basart, Zou, Mazeika, Song, and Steinhardt]{Hendrycks2020}
Dan Hendrycks, Collin Burns, Steven Basart, Andy Zou, Mantas Mazeika, Dawn Song, and Jacob Steinhardt.
\newblock Measuring massive multitask language understanding.
\newblock \emph{arXiv preprint arXiv:2009.03300}, 2021.

\bibitem[Hu et~al.(2024)Hu, Huang, Ilhan, Tekin, Liu, Kompella, and Liu]{Hu2024}
Sihao Hu, Tiansheng Huang, Fatih Ilhan, Selim Tekin, Gaowen Liu, Ramana Kompella, and Ling Liu.
\newblock Pok{\'e}{LLM}on: A human-parity agent for {P}ok{\'e}mon battles with large language models.
\newblock \emph{arXiv preprint arXiv:2402.01118}, 2024.

\bibitem[Huang et~al.(2024)Huang, Liu, Chen, Wang, Wang, Lian, Wang, and Chen]{Huang2024}
Xu~Huang, Weiwen Liu, Xiaolong Chen, Xingmei Wang, Hao Wang, Defu Lian, Yasheng Wang, and Enhong Chen.
\newblock Understanding the planning of {LLM} agents: A survey.
\newblock \emph{arXiv preprint arXiv:2402.02716}, 2024.

\bibitem[Kosinski(2023)]{Kosinski2023}
Michal Kosinski.
\newblock Theory of mind may have spontaneously emerged in large language models.
\newblock \emph{arXiv preprint arXiv:2302.02083}, 2023.

\bibitem[{Meta AI}(2024)]{Meta2024}
{Meta AI}.
\newblock The {L}lama 3 herd of models.
\newblock \emph{arXiv preprint arXiv:2407.21783}, 2024.

\bibitem[{OpenAI}(2023)]{OpenAI2023}
{OpenAI}.
\newblock {GPT}-4 technical report.
\newblock \emph{arXiv preprint arXiv:2303.08774}, 2023.

\bibitem[Ruoss et~al.(2024)Ruoss, Del{\'e}tang, Medapati, Grau-Moya, Wenliang, Catt, Reid, and Genewein]{Ruoss2024}
Anian Ruoss, Gr{\'e}goire Del{\'e}tang, Sourabh Medapati, Jordi Grau-Moya, Li~Ke Wenliang, Elliot Catt, John Reid, and Tim Genewein.
\newblock Grandmaster-level chess without search.
\newblock \emph{arXiv preprint arXiv:2402.04494}, 2024.

\bibitem[Silver et~al.(2016)Silver, Huang, Maddison, Guez, Sifre, Van Den~Driessche, et~al.]{Silver2016}
David Silver, Aja Huang, Chris~J Maddison, Arthur Guez, Laurent Sifre, George Van Den~Driessche, et~al.
\newblock Mastering the game of {G}o with deep neural networks and tree search.
\newblock \emph{Nature}, 529:\penalty0 484--489, 2016.

\bibitem[Talmor et~al.(2019)Talmor, Herzig, Lourie, and Berant]{Talmor2018}
Alon Talmor, Jonathan Herzig, Nicholas Lourie, and Jonathan Berant.
\newblock {CommonsenseQA}: A question answering challenge targeting world knowledge.
\newblock In \emph{Proceedings of NAACL}, 2019.

\bibitem[Tubaishat et~al.(2023)Tubaishat, Anwar, Al-Obeidat, Shah, and Razzaq]{Tubaishat2023}
Abdallah Tubaishat, Sajid Anwar, Feras~N. Al-Obeidat, Babar Shah, and Muhammad~Saad Razzaq.
\newblock An artificially intelligent ludo player.
\newblock In \emph{2023 7th IEEE Congress on Information Science and Technology (CiSt)}. IEEE, 2023.

\bibitem[Wang et~al.(2019)Wang, Singh, Michael, Hill, Levy, and Bowman]{Wang2018}
Alex Wang, Amanpreet Singh, Julian Michael, Felix Hill, Omer Levy, and Samuel Bowman.
\newblock {GLUE}: A multi-task benchmark and analysis platform for natural language understanding.
\newblock In \emph{Proceedings of ICLR}, 2019.

\bibitem[Wei et~al.(2022)Wei, Wang, Schuurmans, Bosma, Ichter, Xia, Chi, Le, and Zhou]{Wei2022}
Jason Wei, Xuezhi Wang, Dale Schuurmans, Maarten Bosma, Brian Ichter, Fei Xia, Ed~Chi, Quoc Le, and Denny Zhou.
\newblock Chain-of-thought prompting elicits reasoning in large language models.
\newblock In \emph{Advances in Neural Information Processing Systems (NeurIPS)}, 2022.

\bibitem[Xu et~al.(2024)Xu, Yu, Fang, Wang, and Wu]{Xu2024}
Zelai Xu, Chao Yu, Fei Fang, Yu~Wang, and Yi~Wu.
\newblock Language agents with reinforcement learning for strategic play in the {W}erewolf game.
\newblock \emph{arXiv preprint arXiv:2310.18940}, 2024.

\bibitem[Yao et~al.(2023{\natexlab{a}})Yao, Yu, Zhao, Shafran, Griffiths, Cao, and Narasimhan]{Yao2023}
Shunyu Yao, Dian Yu, Jeffrey Zhao, Izhak Shafran, Thomas~L Griffiths, Yuan Cao, and Karthik Narasimhan.
\newblock Tree of thoughts: Deliberate problem solving with large language models.
\newblock \emph{arXiv preprint arXiv:2305.10601}, 2023{\natexlab{a}}.

\bibitem[Yao et~al.(2023{\natexlab{b}})Yao, Zhao, Yu, Du, Shafran, Narasimhan, and Cao]{Yao2023react}
Shunyu Yao, Jeffrey Zhao, Dian Yu, Nan Du, Izhak Shafran, Karthik Narasimhan, and Yuan Cao.
\newblock {ReAct}: Synergizing reasoning and acting in language models.
\newblock In \emph{International Conference on Learning Representations (ICLR)}, 2023{\natexlab{b}}.

\bibitem[Zhuang et~al.(2025)Zhuang, Fang, Choi, Du, Liang, Cao, and Tomlin]{Zhuang2025}
Richard Zhuang, Richard Fang, Gail Choi, Tianjun Du, Felicia Liang, Anson Cao, and Claire Tomlin.
\newblock {PokerBench}: Training large language models to become professional poker players.
\newblock In \emph{Proceedings of AAAI 2025}, 2025.

\end{thebibliography}
\newpage
\appendix

\section{Board Model Specifications}
\label{app:board_model}

Table~\ref{tab:board_model} provides the complete board zone specifications referenced in \S\ref{sec:background}.

\begin{table}[H]
\centering
\caption{Board zone specifications.}
\label{tab:board_model}
\resizebox{\textwidth}{!}{%
\begin{tabular}{lll}
\toprule
\textbf{Zone} & \textbf{Indices} & \textbf{Notes} \\
\midrule
Main Board      & 0\ldots51 (52 squares)       & Circular, shared by all players \\
Base (off-board)& $-1$                         & Pieces start here; leave only on dice$=6$ \\
Home Path P0    & 52\ldots57 (home\_end$=57$)  & Private; entered after one full lap \\
Home Path P1    & 58\ldots63 (home\_end$=63$)  & Private \\
Home Path P2    & 64\ldots69 (home\_end$=69$)  & Private \\
Home Path P3    & 70\ldots75 (home\_end$=75$)  & Private \\
Safe Squares    & \{0, 8, 13, 21, 26, 34, 39, 47\} & Pieces cannot be captured here \\
\bottomrule
\end{tabular}%
}
\end{table}
\subsection{Movement and Capture Rules}

Table~\ref{tab:rules} summarizes the exact movement and capture rules as implemented in \texttt{LudoBench}.

\begin{table}[H]
\centering
\caption{Movement and capture rules implemented in \textsc{LudoBench}.}
\label{tab:rules}
\resizebox{\textwidth}{!}{%
\begin{tabular}{p{0.32\textwidth} p{0.62\textwidth}}
\toprule
\textbf{Rule} & \textbf{Behavior} \\
\midrule
Dice range            & 1 to 6 \\[3pt]
Leave base            & Only on dice$=6$; lands on player start square \\[3pt]
Extra turn            & Only on dice$=6$ (NOT for capture, NOT for reaching home) \\[3pt]
Home entry            & Exact landing required; overshoot is illegal \\[3pt]
Capture               & Only on non-safe main-board squares; captured piece $\rightarrow$ base ($-1$) \\[3pt]
Same-player stacking  & Blocked on all squares (exception: \texttt{home\_end}) \\[3pt]
Win condition         & All 4 pieces reach player \texttt{home\_end} \\[3pt]
No-legal-move (non-6) & Turn passes to next player \\[3pt]
No-legal-move (6)     & Same player continues (extra-turn rule applies) \\
\bottomrule
\end{tabular}%
}
\end{table}
\newpage
\section{Spot Category Details}
\label{app:categories}

\subsection*{Group 1: Rule Compliance}

\paragraph{\texttt{blocked}.} The board is configured so that one or more of the LLM's pieces cannot legally move (e.g., destination occupied by own piece). The LLM must identify and avoid these blocked pieces, selecting a valid alternative.\textit{Metrics:} \texttt{block\_rate} (fraction choosing a blocked piece among valid outputs), \texttt{llm\_invalid\_rate}.

\paragraph{\texttt{overshoot}.} A piece is positioned near home-end such that the dice roll would overshoot the exact landing requirement. The LLM must recognize that this move is illegal and select an alternative piece. This tests whether the model correctly avoids illegal moves near the end of the game. \textit{Metrics:} \texttt{overshoot\_rate} (fraction attempting an overshoot), \texttt{llm\_invalid\_rate}.

\subsection*{Group 2: Pure Preference}

\paragraph{\texttt{capture}.} A capture move is available with no competing strategic incentive (no simultaneous home-entry or safety option). Tests whether the model exhibits a baseline aggression tendency when capture is the only salient option. \textit{Metrics:} \texttt{capture\_rate}, \texttt{llm\_invalid\_rate}.

\paragraph{\texttt{home\_entry}.} A move entering or progressing along the home path is available without a competing capture opportunity. Tests goal-seeking behavior of whether the model prioritizes advancing toward the win condition. \textit{Metrics:} \texttt{home\_entry\_rate}, \texttt{llm\_invalid\_rate}.

\paragraph{\texttt{safe}.} A move landing on a safe square is available without competing capture or home-entry options. Tests risk aversion and defensive tendency in the absence of offensive opportunity. \textit{Metrics:} \texttt{safe\_rate}, \texttt{llm\_invalid\_rate}.

\paragraph{\texttt{extra\_turn}.} Dice$=$6 scenario where the LLM must choose between bringing out a new piece from base versus advancing an existing piece on the board. No capture or home-entry is involved. Tests development strategy (board presence) versus advancement strategy (piece progress). \textit{Metrics:} \texttt{bring\_out\_rate}, \texttt{move\_existing\_rate}.

\subsection*{Group 3: Strategic Tradeoff}

\paragraph{\texttt{capture\_vs\_home}.} One piece can capture an opponent; another can enter or advance on the home path. Forces a direct choice between short-term aggression (disrupting an opponent) and long-term goal progress (advancing toward victory). \textit{Metrics:} \texttt{capture\_rate}, \texttt{home\_entry\_rate}.

\paragraph{\texttt{capture\_vs\_home\_finish}.} One piece can capture; another can land on the exact home-end position, completing that piece's journey. The most extreme aggression-vs-objective test: choosing capture here means forgoing an immediate step toward the win condition. \textit{Metrics:} \texttt{capture\_rate}, \texttt{home\_finish\_rate}.

\paragraph{\texttt{capture\_vs\_openexisting}.} Dice$=$6 scenario: one piece can capture an opponent; another option is to open a new piece from base. Tests tactical aggression versus strategic development. \textit{Metrics:} \texttt{capture\_rate}, \texttt{open\_rate}.

\paragraph{\texttt{capture\_vs\_safe}.} One piece can capture an opponent; another can move to a safe square. Directly quantifies the aggression-defense spectrum, revealing the model's risk tolerance when both options are available. \textit{Metrics:} \texttt{capture\_rate}, \texttt{safe\_rate}.

\paragraph{\texttt{safe\_vs\_openexisting}.} Dice$=$6 scenario: one piece can move to a safe square; another option is to open a new piece from base. Tests whether the model prioritizes protecting existing pieces or developing new ones. \textit{Metrics:} \texttt{safe\_rate}, \texttt{open\_rate}.

\subsection*{Group 4: History-Conditioned}

\paragraph{\texttt{grudge\_paired}.} Each scenario presents the \emph{same} board state twice: once with a grudge narrative (e.g., ``Player~2 captured your piece last turn despite having a safer option'') and once with neutral history. It tests whether LLMs exhibit history-sensitive decision shifts and retaliation tendencies, probing a fundamental question about context-dependent behavior in deployed systems. \textit{Metrics:} \texttt{change\_rate}, \texttt{retaliation\_grudge\_rate}, \texttt{retaliation\_noconflict\_rate}.
% \newpage
\section{Metrics Definitions}
\label{app:metrics}

All behavioral rates are computed exclusively on \emph{valid} LLM outputs i.e., outputs where the selected piece index corresponded to a legal move. The sole exception is \texttt{llm\_invalid\_rate}, which is computed on \emph{all} outputs before heuristic fallback. This distinction is critical: it separates the question of ``can the model follow the rules?'' from ``what does the model prefer when it does follow the rules?''

\begin{table*}[h!]
\centering
\small
\caption{Metrics definitions. All behavioral rates computed on valid outputs only; \texttt{llm\_invalid\_rate} computed on all outputs pre-fallback.}
\label{tab:metrics}
\setlength{\tabcolsep}{4pt}
\renewcommand{\arraystretch}{1.15}
\begin{tabular}{@{}llp{7.5cm}@{}}
\toprule
\textbf{Group} & \textbf{Metric} & \textbf{Definition} \\
\midrule
Reliability & \texttt{llm\_invalid\_rate} & Fraction of all spots where the LLM's raw output was not a legal move. Computed \textit{before} heuristic fallback. \\
\midrule
\multirow{7}{*}{Behavioral} 
  & \texttt{capture\_rate} & Fraction choosing the capture move. \\
  & \texttt{safe\_rate} & Fraction choosing the safe-square move. \\
  & \texttt{home\_entry\_rate} & Fraction choosing home path entry or progress. \\
  & \texttt{home\_finish\_rate} & Fraction choosing to finish a piece at home\_end. \\
  & \texttt{open\_rate} & Fraction choosing to bring a base piece into play (dice$=$6). \\
  & \texttt{bring\_out\_rate} & Fraction choosing to bring out a new piece on dice$=$6. \\
  & \texttt{move\_existing\_rate} & Fraction choosing to move an already-active piece on dice$=$6. \\
\midrule
\multirow{2}{*}{Rule Violation}
  & \texttt{block\_rate} & Fraction that chose a self-blocked piece. \\
  & \texttt{overshoot\_rate} & Fraction that attempted an overshoot move past home\_end. \\
\midrule
\multirow{3}{*}{Grudge}
  & \texttt{change\_rate} & Fraction of pairs where the choice differed between grudge and neutral on the same board. \\
  & \texttt{retaliation\_grudge\_rate} & Fraction choosing retaliatory capture under grudge framing. \\
  & \texttt{retaliation\_noconflict\_rate} & Fraction choosing the same retaliatory capture under neutral framing. \\
\bottomrule
\end{tabular}
\end{table*}
\subsection*{Derived Quantities}

The \textbf{grudge effect} is defined as the difference $\Delta = \texttt{retaliation\_grudge\_rate} - \texttt{retaliation\_noconflict\_rate}$. A positive $\Delta$ indicates that grudge framing increases retaliatory behavior; a negative $\Delta$ indicates suppression. \textbf{Transition counts} track the specific types of behavioral shifts between paired conditions (e.g., safe$\to$capture, other$\to$capture), providing qualitative insight beyond the aggregate change\_rate.

\begin{table*}[h!]
\centering
\small
\begin{tabular}{@{}lcccccc|ccc@{}}
\toprule
\textbf{Metric} & \textbf{QP} & \textbf{Q7B} & \textbf{DS} & \textbf{Haiku} & \textbf{Scout} & \textbf{Gemma} & \textbf{GT} & \textbf{Heur.} & \textbf{Rand.} \\
\midrule
\multicolumn{10}{@{}l}{\textit{Pure Preference (none persona)}} \\
\quad capture\_rate        & 0.50 & 0.55 & 0.15 & 0.53 & 0.22 & 0.29 & 0.88 & 1.00 & 0.45 \\
\quad safe\_rate           & 0.28 & 0.47 & 0.12 & 0.75 & 0.05 & 0.15 & 0.88 & 0.75 & 0.53 \\
\quad home\_entry\_rate    & 0.42 & 0.79 & 0.17 & 0.82 & 0.14 & 0.16 & 0.13 & 1.00 & 0.50 \\
\quad bring\_out\_rate     & 0.30 & 0.60 & 0.88 & 0.05 & 0.70 & 0.82 & 1.00 & 1.00 & 0.78 \\
\quad move\_existing\_rate & 0.70 & 0.40 & 0.12 & 0.95 & 0.30 & 0.18 & 0.00 & 0.00 & 0.22 \\
\midrule
\multicolumn{10}{@{}l}{\textit{Key Tradeoff Rates (none persona)}} \\
\quad home\_finish\_rate (cvf) & 0.80 & 1.00 & 0.00 & 0.88 & 0.05 & 0.00 & 1.00 & 0.00 & 0.53 \\
\quad cvs capture\_rate    & 0.42 & 0.61 & 0.10 & 0.66 & 0.29 & 0.34 & 0.98 & 1.00 & 0.63 \\
\quad cvs safe\_rate       & 0.57 & 0.39 & 0.90 & 0.34 & 0.71 & 0.66 & 0.03 & 0.00 & 0.38 \\
\bottomrule
\end{tabular}
\caption{Full behavioral profiles under the neutral persona. GT = Game-Theory Agent, Heur.\ = Heuristic Agent, Rand.\ = Random Agent. All baselines are deterministic and persona-invariant.}
\label{tab:profiles}
\end{table*}

\begin{table*}[h!]
\centering
\small
\begin{tabular}{@{}llcccccc|ccc@{}}
\toprule
\textbf{Category} & \textbf{Metric} & \textbf{QP} & \textbf{Q7B} & \textbf{DS} & \textbf{Haiku} & \textbf{Scout} & \textbf{Gemma} & \textbf{GT} & \textbf{Heur.} & \textbf{Rand.} \\
\midrule
\multirow{2}{*}{\texttt{cvs\_safe}}
  & capture & 0.42 & 0.61 & 0.10 & 0.66 & 0.29 & 0.34 & 0.98 & 1.00 & 0.63 \\
  & safe    & 0.57 & 0.39 & 0.90 & 0.34 & 0.71 & 0.66 & 0.03 & 0.00 & 0.38 \\
\midrule
\multirow{2}{*}{\texttt{cvs\_home}}
  & capture    & 0.57 & 0.15 & 0.72 & 0.20 & 0.78 & 0.83 & 0.98 & 1.00 & 0.35 \\
  & home\_entry & 0.42 & 0.85 & 0.28 & 0.80 & 0.22 & 0.17 & 0.03 & 0.00 & 0.65 \\
\midrule
\multirow{2}{*}{\texttt{cvs\_finish}}
  & capture      & 0.20 & 0.00 & 1.00 & 0.12 & 0.95 & 1.00 & 0.00 & 1.00 & 0.48 \\
  & home\_finish & 0.80 & 1.00 & 0.00 & 0.88 & 0.05 & 0.00 & 1.00 & 0.00 & 0.53 \\
\midrule
\multirow{2}{*}{\texttt{cvs\_open}}
  & capture & 0.62 & 0.50 & 0.23 & 0.93 & 0.27 & 0.31 & 0.68 & 1.00 & 0.23 \\
  & open    & 0.38 & 0.50 & 0.78 & 0.07 & 0.73 & 0.69 & 0.33 & 0.00 & 0.78 \\
\bottomrule
\end{tabular}
\caption{Tradeoff category decision rates under the neutral persona. cvs = capture\_vs. GT = Game-Theory Agent, Heur.\ = Heuristic Agent, Rand.\ = Random Agent.}
\label{tab:tradeoffs}
\end{table*}

\section{Persona Effect Details}
\label{app:persona_details}
\begin{table}[t]
\centering
\small
\begin{tabular}{@{}lcccccc@{}}
\toprule
\textbf{Metric} & \textbf{QP} & \textbf{Q7B} & \textbf{DS} & \textbf{Haiku} & \textbf{Scout} & \textbf{Gemma} \\
\midrule
change\_rate             & \textbf{0.33} & 0.00 & 0.20 & 0.05 & 0.06 & 0.03 \\
ret.\ grudge             & 0.80 & 0.57 & 0.78 & 0.31 & 0.86 & 0.88 \\
ret.\ noconflict         & 0.68 & 0.57 & 0.73 & 0.33 & 0.83 & 0.85 \\
\midrule
grudge effect ($\Delta$) & +0.12 & 0.00 & +0.05 & $-$0.03 & +0.03 & +0.03 \\
\bottomrule
\end{tabular}
\caption{Grudge sensitivity under the neutral persona. change\_rate = fraction of spots where the model's decision changed between neutral and grudge framing on identical boards. The GT agent scores 0\% on all metrics.}
\label{tab:grudge}
\end{table}

Table~\ref{tab:persona_effects} provides the full per-category persona breakdown referenced in \S\ref{sec:persona_effects}.

\begin{table*}[h]
\centering
\scriptsize
\setlength{\tabcolsep}{2.5pt}
\caption{Persona effects on key metrics across all six models. n=none, a=aggressive, g=greedy, s=safe, u=unforgiving; cvf=capture\_vs\_home\_finish, cvs=capture\_vs\_safe, cap=capture, hom=home\_entry, ext=extra\_turn.}
\label{tab:persona_effects}
\vspace{0.3em}
\resizebox{\textwidth}{!}{%
\begin{tabular}{@{}ll|ccccc|ccccc|ccccc|ccccc|ccccc|ccccc@{}}
\toprule
& & \multicolumn{5}{c|}{\textbf{QP}} & \multicolumn{5}{c|}{\textbf{Q7B}} & \multicolumn{5}{c|}{\textbf{DS}} & \multicolumn{5}{c|}{\textbf{Haiku}} & \multicolumn{5}{c|}{\textbf{Scout}} & \multicolumn{5}{c}{\textbf{Gemma}} \\
\textbf{Cat.} & \textbf{Met.} & \textit{n} & \textit{a} & \textit{g} & \textit{s} & \textit{u} & \textit{n} & \textit{a} & \textit{g} & \textit{s} & \textit{u} & \textit{n} & \textit{a} & \textit{g} & \textit{s} & \textit{u} & \textit{n} & \textit{a} & \textit{g} & \textit{s} & \textit{u} & \textit{n} & \textit{a} & \textit{g} & \textit{s} & \textit{u} & \textit{n} & \textit{a} & \textit{g} & \textit{s} & \textit{u} \\
\midrule
cvf & cap  & .20 & .93 & .12 & .62 & .57  & .00 & .14 & .06 & .00 & .16  & 1.0 & 1.0 & 1.0 & 1.0 & 1.0  & .12 & .05 & .15 & .03 & .25  & .95 & .97 & 1.0 & .92 & 1.0  & 1.0 & 1.0 & 1.0 & .94 & .95 \\
cvf & fin  & .80 & .07 & .88 & .38 & .42  & 1.0 & .86 & .94 & 1.0 & .84  & .00 & .00 & .00 & .00 & .00  & .88 & .95 & .85 & .97 & .75  & .05 & .03 & .00 & .08 & .00  & .00 & .00 & .00 & .06 & .05 \\
cvs & cap  & .42 & .15 & .53 & .40 & .15  & .61 & .71 & .80 & .63 & .39  & .10 & .20 & .23 & .15 & .25  & .66 & .84 & .75 & .88 & .72  & .29 & .19 & .21 & .09 & .19  & .34 & .14 & .17 & .11 & .17 \\
cvs & safe & .57 & .85 & .47 & .60 & .85  & .39 & .29 & .20 & .37 & .61  & .90 & .80 & .77 & .85 & .75  & .34 & .16 & .25 & .12 & .28  & .71 & .81 & .79 & .91 & .81  & .66 & .86 & .83 & .89 & .83 \\
cap & cap  & .50 & .28 & .57 & .38 & .33  & .55 & .67 & .50 & .68 & .65  & .15 & .17 & .30 & .07 & .17  & .53 & .90 & .78 & .85 & .74  & .22 & .16 & .29 & .17 & .19  & .29 & .16 & .24 & .21 & .16 \\
hom & h\_e & .42 & .33 & .47 & .33 & .30  & .79 & .69 & .77 & .80 & .69  & .17 & .15 & .28 & .23 & .25  & .82 & .97 & .93 & .97 & .80  & .14 & .00 & .05 & .10 & .05  & .16 & .05 & .03 & .05 & .05 \\
ext & b\_o & .30 & .20 & .28 & .30 & .28  & .60 & .30 & .42 & .42 & .38  & .88 & .55 & .90 & .90 & .72  & .05 & .05 & .03 & .05 & .05  & .70 & .33 & .47 & .51 & .57  & .82 & .51 & .58 & .51 & .64 \\
\bottomrule
\end{tabular}%
}
\end{table*}
\newpage
\section{Persona Prompts}
\label{app:personas}

Each persona is injected into the LLM prompt as a dedicated \texttt{PERSONA} block (see Appendix~\ref{app:prompt} for exact placement). The persona text appears after the game rules and before the current game state, framing the model's strategic orientation before it processes board information.

\subsection*{none (neutral)}

No persona block is included. The model receives the standard game prompt only, establishing a baseline behavioral profile uninfluenced by instructional framing. All metrics under the \textit{none} condition serve as the reference point for measuring persona effects.

\subsection*{aggressive}

\begin{quote}
\small\ttfamily
--------------------\\
PERSONA\\
--------------------\\[4pt]
You must play with this persona style:\\
You are an aggressive Ludo player. Prioritize capturing opponent pieces whenever possible. Attacking opponents is more important than protecting your own pieces or advancing toward home.
\end{quote}

\noindent\textbf{Expected effect:} Elevated \texttt{capture\_rate} across all tradeoff categories. Reduced \texttt{safe\_rate} and \texttt{home\_entry\_rate}.

\subsection*{greedy}

\begin{quote}
\small\ttfamily
--------------------\\
PERSONA\\
--------------------\\[4pt]
You must play with this persona style:\\
You are a greedy Ludo player focused on winning. Prioritize advancing your own pieces toward home above all else. Getting pieces to the finish is more important than capturing opponents.
\end{quote}

\noindent\textbf{Expected effect:} Elevated \texttt{home\_entry\_rate} and \texttt{home\_finish\_rate}. Reduced \texttt{capture\_rate}.

\subsection*{safe}

\begin{quote}
\small\ttfamily
--------------------\\
PERSONA\\
--------------------\\[4pt]
You must play with this persona style:\\
You are a cautious Ludo player. Prioritize moving to safe squares and avoiding risky positions. Protecting your pieces from capture is more important than aggressive play.
\end{quote}

\noindent\textbf{Expected effect:} Elevated \texttt{safe\_rate}. Reduced \texttt{capture\_rate} across tradeoff categories.

\subsection*{unforgiving}

\begin{quote}
\small\ttfamily
--------------------\\
PERSONA\\
--------------------\\[4pt]
You must play with this persona style:\\
You are an unforgiving Ludo player. If an opponent has captured your piece, prioritize retaliating by capturing their pieces in return. Do not let attacks go unpunished.
\end{quote}

\noindent\textbf{Expected effect:} Elevated \texttt{retaliation\_grudge\_rate} and \texttt{change\_rate} in grudge scenarios. Higher \texttt{capture\_rate} specifically targeting prior aggressors.

\newpage
\section{Heuristic Agent Scoring Formula}
\label{app:heuristic}

The HeuristicAgent provides a deterministic, reproducible baseline for comparing LLM decisions. For each legal piece move, it computes a numeric score and selects the piece with the highest value. The complete scoring logic, extracted from \texttt{agents.py}, is as follows.

\subsection*{Scoring Rules}

\begin{table}[h]
\centering
\small
\begin{tabular}{@{}p{3.0cm}p{2.5cm}p{1.8cm}@{}}
\toprule
\textbf{piece State} & \textbf{Score Formula} & \textbf{Range} \\
\midrule
Base ($\text{pos} = -1$) & Fixed value & 50 \\[2pt]
Main board, stays on main & $\text{rel\_pos} + \text{dice}$ & 0--51 \\[2pt]
Main board, enters home & $52 + \text{home\_step}$ & 53+ \\[2pt]
Already in home path & $52 + (\text{new\_pos} - \text{home\_start} + 1)$ & Increasing \\[2pt]
Capture available & $+100$ bonus (non-safe square, opponent present) & Additive \\[2pt]
Safe square destination & $+20$ bonus (dest.\ is safe) & Additive \\
\bottomrule
\end{tabular}
\caption{HeuristicAgent scoring formula. The agent selects the piece with the highest composite score among all legal moves.}
\label{tab:app_heuristic}
\end{table}

\subsection*{Resulting Priority Order}

The scoring formula produces the following implicit priority hierarchy:

\begin{enumerate}[leftmargin=*,itemsep=2pt]
    \item \textbf{Capture} ($+100$): Any legal capture on a non-safe square dominates all other move types. The heuristic always captures when possible.
    \item \textbf{Home path progress} ($52+$): Entering or advancing in the home path is valued above all non-capture main-board moves.
    \item \textbf{Leave base} ($50$): Bringing a piece out of base is valued slightly below home entry but above most main-board advancement.
    \item \textbf{Safe square} ($+20$): Landing on a safe square provides a moderate defensive bonus, but does not override capture or home-entry incentives.
    \item \textbf{Main-board advancement} ($0\text{--}51$): Basic forward progress, with higher scores for positions further from start.
\end{enumerate}
\newpage
\section{LLM Prompt Template}
\label{app:prompt}

The following is the complete prompt template provided to the LLM agent, reproduced from \texttt{llm\_agent.py}. Fields in \texttt{[CAPS]} are populated dynamically from the spot configuration. In spot evaluation mode, the legal moves section is \emph{omitted} (\texttt{include\_legal\_moves\_in\_prompt=False}).

\subsection*{Full Prompt}

\begin{small}
\begin{verbatim}
You are an AI agent playing Ludo as Player
[PLAYER_ID]. Your job is to choose exactly ONE
of your token indices.

IMPORTANT:
- You must output ONLY a token index (0,1,2,3)
  followed by " | " and a one-line reason.
- Do NOT output board positions or text without
  the token index.
- The token index MUST refer to one of your own
  4 tokens (0-3).
- If you output anything else, it is invalid.


--------------------
BOARD & POSITION SYSTEM

--------------------
1. Main circular board: 52 squares (0-51).
2. Each player has a fixed START square.
3. Tokens move forward relative to START.
4. After one full lap (52 steps), tokens enter
   that player's HOME PATH.
5. Each player has a UNIQUE HOME PATH (>= 52).
6. Final home position is HOME_END.
7. Tokens must land EXACTLY on HOME_END.
8. Overshooting HOME_END is illegal.


--------------------
TOKEN STATES

--------------------
- Position = -1  : Token is in base
- Position 0-51  : Token is on main board
- Position >= 52 : Token is in home path
- HOME_END       : Token has finished


--------------------
GAME RULES
--------------------
1.  All tokens start in base (-1).
2.  Leave base ONLY on dice = 6.
3.  Leaving base places token at START square.
4.  Tokens move forward by dice value.
5.  No stacking (one token per square).
6.  CAPTURE: land on opponent on non-safe square
    -> opponent sent to base (-1).
7.  Captures NEVER happen on safe squares.
8.  Safe squares protect tokens from capture.
9.  Rolling 6 grants an extra turn.
10. No legal move -> turn skipped.
11. First to move ALL tokens to HOME_END wins.
12. Home paths are private to each player.

--------------------
CURRENT GAME STATE
--------------------
Number of players: [NUM_PLAYERS]
Active player ids: [PLAYER_IDS]
Dice rolled: [DICE]

Your token positions (Player [PLAYER_ID]):
[YOUR_TOKENS]

Other players' token positions:
[OTHER_PLAYERS_TOKENS]

Your start square: [START_POS]
Your home path: [HOME_START] to [HOME_END]
Safe squares: [SAFE_SQUARES]
Player path ranges: [PLAYER_INFO]

[PERSONA BLOCK — if persona active]
[HISTORY BLOCK — if grudge scenario]

Choose the BEST legal move to win.

--------------------
OUTPUT FORMAT (STRICT)
--------------------
<int> | <one line reason>
\end{verbatim}
\end{small}

\subsection*{Dynamic Blocks}

Two optional blocks are injected depending on experimental condition:

\paragraph{Persona Block.} When a persona is active (all conditions except \textit{none}), the following block is inserted after the game state:

\begin{small}
\begin{verbatim}
--------------------
PERSONA
--------------------
You must play with this persona style:
[PERSONA_TEXT]
\end{verbatim}
\end{small}

\paragraph{History Block.} For \texttt{grudge\_paired} scenarios, a history block is injected providing narrative context about prior game events:

\begin{small}
\begin{verbatim}
--------------------
HISTORY / CONTEXT
--------------------
[HISTORY_TEXT]
\end{verbatim}
\end{small}

\needspace{10\baselineskip}
\vspace*{0pt}
% \newpage
\begin{table}[H]
\centering
\small
\renewcommand{\arraystretch}{1.1}
\caption{Dataset summary.\textsuperscript{$\dagger$} Each category's source file follows the pattern \texttt{spots\_<category>.json}. \texttt{grudge\_paired} contains 40 boards $\times$ 2 conditions (neutral \texttt{\_a} and grudge \texttt{\_b}), yielding 80 entries from 40 unique boards.}
\label{tab:dataset_summary}
\vspace{0.3em}
\begin{tabular}{@{}lccc@{}}
\toprule
\textbf{Category} & \textbf{Entries} & \textbf{Unique Boards} & \textbf{History} \\
\midrule
\texttt{blocked}                & 40 & 40 & No  \\
\texttt{capture}                & 40 & 40 & No  \\
\texttt{capture\_vs\_home}       & 40 & 40 & No  \\
\texttt{capture\_vs\_home\_finish}& 40 & 40 & No  \\
\texttt{capture\_vs\_openexist.} & 40 & 40 & No  \\
\texttt{capture\_vs\_safe}       & 40 & 40 & No  \\
\texttt{extra\_turn}             & 40 & 40 & No  \\
\texttt{grudge\_paired}          & 80 & 40 & Yes \\
\texttt{home\_entry}             & 40 & 40 & No  \\
\texttt{overshoot}              & 40 & 40 & No  \\
\texttt{safe}                   & 40 & 40 & No  \\
\texttt{safe\_vs\_openexist.}    & 40 & 40 & No  \\
\midrule
\textbf{Total}                  & \textbf{520} & \textbf{480} & \\
\bottomrule
\end{tabular}
\end{table}

\vspace{0.5em}

\noindent Table~\ref{tab:dataset_distribution} shows the distribution across game configurations and LLM player assignments, confirming balanced coverage.

\begin{table*}[!h]
\centering
\renewcommand{\arraystretch}{1.7}
\begin{tabular}{@{}lccc@{}}
\toprule
& \multicolumn{3}{c}{\textbf{Player Count}} \\
\textbf{Category} & \textbf{2P} & \textbf{3P} & \textbf{4P} \\
\midrule
\texttt{blocked}                & 14 & 13 & 13 \\
\texttt{capture}                & 14 & 13 & 13 \\
\texttt{capture\_vs\_home}       & 14 & 13 & 13 \\
\texttt{capture\_vs\_home\_finish}& 14 & 13 & 13 \\
\texttt{capture\_vs\_openexist.} & 14 & 13 & 13 \\
\texttt{capture\_vs\_safe}       & 14 & 13 & 13 \\
\texttt{extra\_turn}             & 14 & 13 & 13 \\
\texttt{grudge\_paired}$^\dagger$ & 28 & 26 & 26 \\
\texttt{home\_entry}             & 14 & 13 & 13 \\
\texttt{overshoot}              & 14 & 13 & 13 \\
\texttt{safe}                   & 14 & 13 & 13 \\
\texttt{safe\_vs\_openexisting}  & 14 & 13 & 13 \\
\bottomrule
\end{tabular}
\caption{Distribution of spots across player configurations (2-player, 3-player, 4-player). $^\dagger$\texttt{grudge\_paired} counts are doubled due to \texttt{\_a}/\texttt{\_b} pairing. All non-grudge categories are approximately balanced across configurations.}
\label{tab:dataset_distribution}
\end{table*}

\vspace{0.5em}

\noindent Each spot entry contains the fields shown in Table~\ref{tab:spot_schema}. Figure~\ref{fig:spot_example} provides a representative example.

\begin{table*}[!h]
\centering
\renewcommand{\arraystretch}{1.6}
\begin{tabular}{@{}p{3.2cm}p{2.5cm}p{8.5cm}@{}}
\toprule
\textbf{Field} & \textbf{Type} & \textbf{Description} \\
\midrule
\texttt{id} & string & Unique identifier (e.g., \texttt{cvs\_2p\_001}, \texttt{grudge\_pair1001\_a}) \\
\texttt{scenario} & string & Category label matching one of the 12 evaluation types \\
\texttt{players} & list[int] & Active player IDs in this game (e.g., \texttt{[0, 1]} for 2-player) \\
\texttt{llm\_player\_id} & int & Which player index the LLM controls \\
\texttt{current\_player} & int & Whose turn it is (always equals \texttt{llm\_player\_id}) \\
\texttt{dice} & int & Dice roll value for this decision (1--6) \\
\texttt{tokens} & dict & Token positions for all players; keys are player IDs, values are lists of 4 positions ($-1$ = base, $0$--$51$ = main board, $\geq 52$ = home path) \\
\texttt{note} & string & Human-readable description of the intended decision \\
\texttt{history\_text} & string & \textit{(grudge\_paired only)} Narrative context: neutral or grudge framing \\
\bottomrule
\end{tabular}
\caption{JSON schema for each spot entry. The \texttt{history\_text} field is present only in \texttt{grudge\_paired} scenarios.}
\label{tab:spot_schema}
\end{table*}

\begin{figure*}[h!]
\centering
\small
\begin{minipage}[t]{0.47\textwidth}
\begin{verbatim}
{
  "id": "cvs_2p_001",
  "scenario": "capture_vs_safe",
  "players": [0, 1],
  "llm_player_id": 1,
  "current_player": 1,
  "dice": 6,
  "tokens": {
    "0": [49, -1, -1, -1],
    "1": [43, 41, -1, -1]
  },
  "note": "One capture option and
           one safe-square option."
}
\end{verbatim}
\centerline{(a) \texttt{capture\_vs\_safe} spot}
\end{minipage}
\hfill
\begin{minipage}[t]{0.47\textwidth}
\begin{verbatim}
// Pair _a (neutral):
{
  "id": "grudge_pair1001_a",
  "scenario": "grudge",
  "dice": 5,
  "tokens": {
    "0": [18, 11, -1, -1],
    "1": [23, 16, -1, -1]
  },
  "history_text":
    "No prior conflicts noted."
}
// Pair _b (grudge):
{
  "id": "grudge_pair1001_b",
  // identical tokens & dice
  "history_text":
    "Player 1 captured one of your
     tokens earlier in the game."
}
\end{verbatim}
\centerline{(b) \texttt{grudge\_paired} spot pair}
\end{minipage}
\caption{Example spot entries. (a)~A \texttt{capture\_vs\_safe} scenario: Player~1 (LLM) with dice$=$6 can either capture Player~0's token at square~49 or move to safety. (b)~A grudge pair: identical board state with neutral (\texttt{\_a}) and grudge (\texttt{\_b}) history framing.}
\label{fig:spot_example}

\end{figure*}

\needspace{10\baselineskip}
\vspace*{0pt}
\newpage
\section{Game-Theory Agent}
\label{app:gt_agent}

The GameTheoryAgent provides a principled strategic baseline by computing move values through depth-limited search over the game tree. We describe the 2-player formulation; the multiplayer extension follows analogously.

\subsection*{Recursion Structure}

The agent uses mutual recursion between two node types:

\textbf{Decision nodes} (MAX/MIN): The root player maximizes expected value; the opponent minimizes it. For each legal move $a \in A(s, p, d)$, the agent applies the move to produce a successor state and recurses to a chance node.

\textbf{Chance nodes}: For each dice outcome $d \in \{1, \ldots, 6\}$, the agent recurses to a decision node and averages the results:
\begin{equation}
V_{\text{chance}}(s, p, h) = \frac{1}{6} \sum_{d=1}^{6} V_{\text{dec}}(s, p, d, h)
\end{equation}

\subsection*{Decision Value}

For the root player (MAX):
\begin{equation}
V_{\text{dec}}(s, p, d, h) = \max_{a \in A(s,p,d)} V_{\text{chance}}\big(T(s,p,d,a),\; p',\; h{-}1\big)
\end{equation}

For the opponent (MIN):
\begin{equation}
V_{\text{dec}}(s, p, d, h) = \min_{a \in A(s,p,d)} V_{\text{chance}}\big(T(s,p,d,a),\; p',\; h{-}1\big)
\end{equation}

where $T$ is the transition function, $p' = p$ if $d = 6$ (extra turn), else the other player, and $h$ is remaining depth.

\subsection*{Terminal and Cutoff Values}

Terminal states return $+1$ (root wins) or $-1$ (root loses). At depth cutoff ($h = 0$), a weighted linear evaluator scores the state:
\begin{equation}
V_{\text{eval}} = 0.0025 \cdot \Delta_{\text{prog}} + 0.20 \cdot \Delta_{\text{fin}} - 0.05 \cdot \Delta_{\text{base}} + 0.01 \cdot \Delta_{\text{safe}}
\end{equation}

where each $\Delta$ term is the difference between the root player's and opponent's count (progress along board, finished pieces, pieces in base, pieces on safe squares). Values are clipped to $[-0.999, 0.999]$.

\subsection*{Memoization}

A transposition cache stores previously computed subtrees, keyed by (node type, serialized state, current player, dice, depth). This avoids redundant computation in the exponentially branching game tree.

\subsection*{Multiplayer Extension (3/4 Players)}

For $n > 2$ players, the agent switches to Expectimax-MaxN: each player maximizes their own component of a vector-valued utility. Terminal values assign $+1$ to the winner and $-1/(n{-}1)$ to each loser. The evaluator returns a vector of per-player scores, centered by subtracting the mean. For 2-player games, the multiplayer agent delegates to the exact Expectiminimax implementation.

\subsection*{Complexity and Limitations}

The agent uses a search depth of $h=2$ for all experiments reported in this paper.Without memoization, the search tree grows as $O((6b)^h)$ where $b$ is the average branching factor and $h$ is depth. With caching, practical runtime is substantially reduced. The agent is depth-limited (not globally optimal), uses a hand-tuned linear evaluator, and assumes uniform dice probabilities. Despite these limitations, it provides a substantially stronger baseline than greedy heuristics by explicitly modeling opponent responses and stochastic uncertainty.

\end{document}